\documentclass[10pt,twocolumn,letterpaper]{article}

\usepackage{iccv}
\usepackage{times}
\usepackage{epsfig}
\usepackage{graphicx}
\usepackage{amsmath}
\usepackage{amssymb}

\usepackage[pagebackref=true,breaklinks=true,letterpaper=true,colorlinks,bookmarks=false]{hyperref}

\usepackage{siunitx}
\usepackage{times}
\usepackage{url}
\usepackage{epsfig}
\usepackage{caption}
\usepackage{graphicx}
\usepackage{amsmath,amssymb} 
\usepackage{mathtools}
\usepackage{booktabs}
\usepackage{amsfonts}
\usepackage{upgreek}
\usepackage[mathscr]{euscript}

\usepackage[usenames,dvipsnames,svgnames,table]{xcolor}

\usepackage{capt-of}
\usepackage{dirtytalk}
\usepackage{algorithm}
\usepackage[noend]{algpseudocode}
\usepackage[super]{nth}

\usepackage{multirow}
\usepackage{subfigure}
\usepackage[table]{xcolor}
\usepackage{bm}
\usepackage{subfigure}
\usepackage{balance}
\usepackage{hhline,colortbl}
\usepackage{makecell}
\usepackage{comment}
\usepackage{xspace}
\usepackage{threeparttable}

\usepackage{amsthm}
\usepackage[toc,page]{appendix}
\usepackage{multibib}
\newcites{Supp}{Supplementary References}

\definecolor{Gray}{gray}{0.9}
\definecolor{Gray1}{gray}{0.97}
\graphicspath{{./}{./figures/}}
\DeclareGraphicsExtensions{.pdf,.jpeg,.png,.jpg,.eps}

\definecolor{ao}{rgb}{0.0, 0.5, 0.0}

\def\onedot{\ifx\@let@token.\else.\null\fi\xspace}
\def\eg{\emph{e.g}\onedot} 
\def\eg{\emph{e.g}\onedot} 
\def\ie{\emph{i.e}\onedot} 
 
\def\etc{\emph{etc}\onedot} 
 
\def\etal{\emph{et al}\onedot}
\makeatother

\def\Vec#1{{\boldsymbol{#1}}}
\def\Mat#1{{\boldsymbol{#1}}}

\DeclareMathOperator{\relu}{ReLU}
\DeclareMathOperator{\elu}{ELU}
\DeclareMathOperator{\leakyrelu}{LeakyReLU}
\DeclareMathOperator{\softmax}{Softmax}
\DeclareMathOperator{\bn}{BN}

\DeclareMathOperator{\concat}{Concat}

\DeclareMathOperator{\embed}{Embedding}
\DeclareMathOperator{\gnn}{GNN}
\DeclareMathOperator{\mlp}{MLP}

\newcommand{\RNum}[1]{\lowercase\expandafter{\romannumeral #1\relax}}

\iccvfinalcopy 

\def\iccvPaperID{3520} 

\ificcvfinal\fi

\makeatletter
\newcommand{\printfnsymbol}[1]{%
  \textsuperscript{\@fnsymbol{#1}}%
}

\begin{document}

\title{Feature Correlation Aggregation: on the Path to Better Graph Neural Networks}

\author{Jieming Zhou\\
Australian National University\\
Canberra, Acton 2601, Australia\\
{\tt\small Jieming.Zhou@anu.edu.au}
\and
Tong Zhang\thanks{Corresponding author.}\\
École polytechnique fédérale de Lausanne\\
Rte Cantonale, 1015 Lausanne, Switzerland\\
{\tt\small tong.zhang@epfl.ch}

\and
Pengfei Fang\\
Australian National University\\
Canberra, Acton 2601, Australia\\
{\tt\small Pengfei.Fang@anu.edu.au}

\and
Lars Petersson\\
Data61, CSIRO\\
Canberra, Acton 2601, Australia\\
{\tt\small lars.petersson@data61.csiro.au}

\and
Mehrtash Harandi\\
Monash University\\
Wellington Rd, Clayton VIC 3800, Australia\\
{\tt\small mehrtash.harandi@monash.edu}
}

\maketitle
\ificcvfinal\thispagestyle{empty}\fi

\begin{abstract}

Prior to the introduction of Graph Neural Networks (GNNs), modeling and analyzing irregular data, particularly graphs, was thought to be the Achilles' heel of deep learning. The core concept of GNNs is to find a representation by recursively aggregating the representations of a central node and those of its neighbors. The core concept of GNNs is to find a representation by recursively aggregating the representations of a central node and those of its neighbor, and its success has been demonstrated by many GNNs' designs. However, most of them only focus on using the first-order information between a node and its neighbors. In this paper,  we introduce a central node permutation variant function through a frustratingly simple and innocent-looking modification to the core operation of a GNN, namely the \textbf{F}eature c\textbf{O}rrelation a\textbf{G}gregation (FOG) module which learns the second-order information from feature correlation between a node and its neighbors in the pipeline. By adding FOG into existing variants of GNNs, we empirically verify \footnote{The source code is available at \url{https://github.com/Anonymous/FOG}} this second-order information complements the features generated by original GNNs across a broad set of benchmarks. A tangible boost in performance of the model is observed where the model surpasses previous state-of-the-art results by a significant margin while employing fewer parameters. (\eg, 33.116\% improvement on a real-world molecular dataset using graph convolutional networks).

\end{abstract}

\section{Introduction}
Deep learning, especially in the form of convolutional neural networks (CNNs), has achieved tremendous successes in various machine learning tasks, such as image classification~\cite{imagenet_cvpr09}, object detection~\cite{lin2014microsoft} and machine translation~\cite{vaswani2017attention} \etc However, it remains a big challenge dealing with non-grid or irregular data such as \eg protein-interaction networks, social networks, and knowledge graphs that may best be considered graph-structured data. Due to an abundance of such graph-structured data, graph neural networks (GNNs) have been attracting an increasing level of attention and have successfully been applied to a number of tasks. Consequently, we can see in the literature that much research effort has gone into investigating deep learning architectures and finding powerful representations for such graph-structured data. 



On the path to finding powerful and discriminative representations, exploiting pairwise relationships within graph data and their feature vectors in a principled way has become a pivotal part of the graph learning area. One of the earliest works can be traced back to~\cite{henaff2015deep} in 2015, and the following works including Chebyshev Convolutional Neural Networks (ChebyNets)~\cite{defferrard2016convolutional}, Graph Convolutional Neural Networks (GCNs)~\cite{kipf2016semi}, GraphSAGE~\cite{hamilton2017inductive} and Graph Attention Networks (GATs)~\cite{velivckovic2017graph} demonstrate the importance of leveraging neighborhood information. Although those methods are derived from different perspectives (the spatial and spectral domain), all of them follows message passing scheme where messages between connected nodes are iteratively passed and permutation-invariant aggregation functions (such as mean, summation, and maximum) are employed in each layer to learn a representation for each node or graph.


Recent work~\cite{xu2018powerful} proposes a simple architecture, GIN, which is as powerful as Weisfeiler-Lehman (WL) \cite{leman1968reduction}. Meanwhile, they unveil that a GNN can achieve this when it learns a injective multiset function which should consist a summation aggregator~\cite{xu2018powerful}. A \textit{multiset}, mathematically speaking, denotes a set of feature vectors of a node and its neighbors, while nodes' class and their feature vectors can be repeated. In this scenario, GNNs have mean or maximum aggregator will fail to learn a injective multiset function. However, during training stage, a GNN may still generate indiscriminative features after the sumation aggregator when the network have not converged to an injection function, which will harm the performance. 

Moreover, by gathering information layer by layer, all the nodes in a \textit{multiset} tend to have similar representations, which is also known as over-smoothing. This is also the reason why these types of works are favored by assortative graphs such as citation networks but degenerate significantly in disassortative graphs where the nodes of the same class share a high structural similarity but are far apart from each other~\cite{ribeiro2017struc2vec}. This phenomenon is consistent with the most recent paper~\cite{dwivedi2020benchmarking} which thoroughly demonstrates the performance of different architectures on various benchmarks.

\begin{figure}[!ht]
  \centering
  \includegraphics[width=8cm,keepaspectratio]{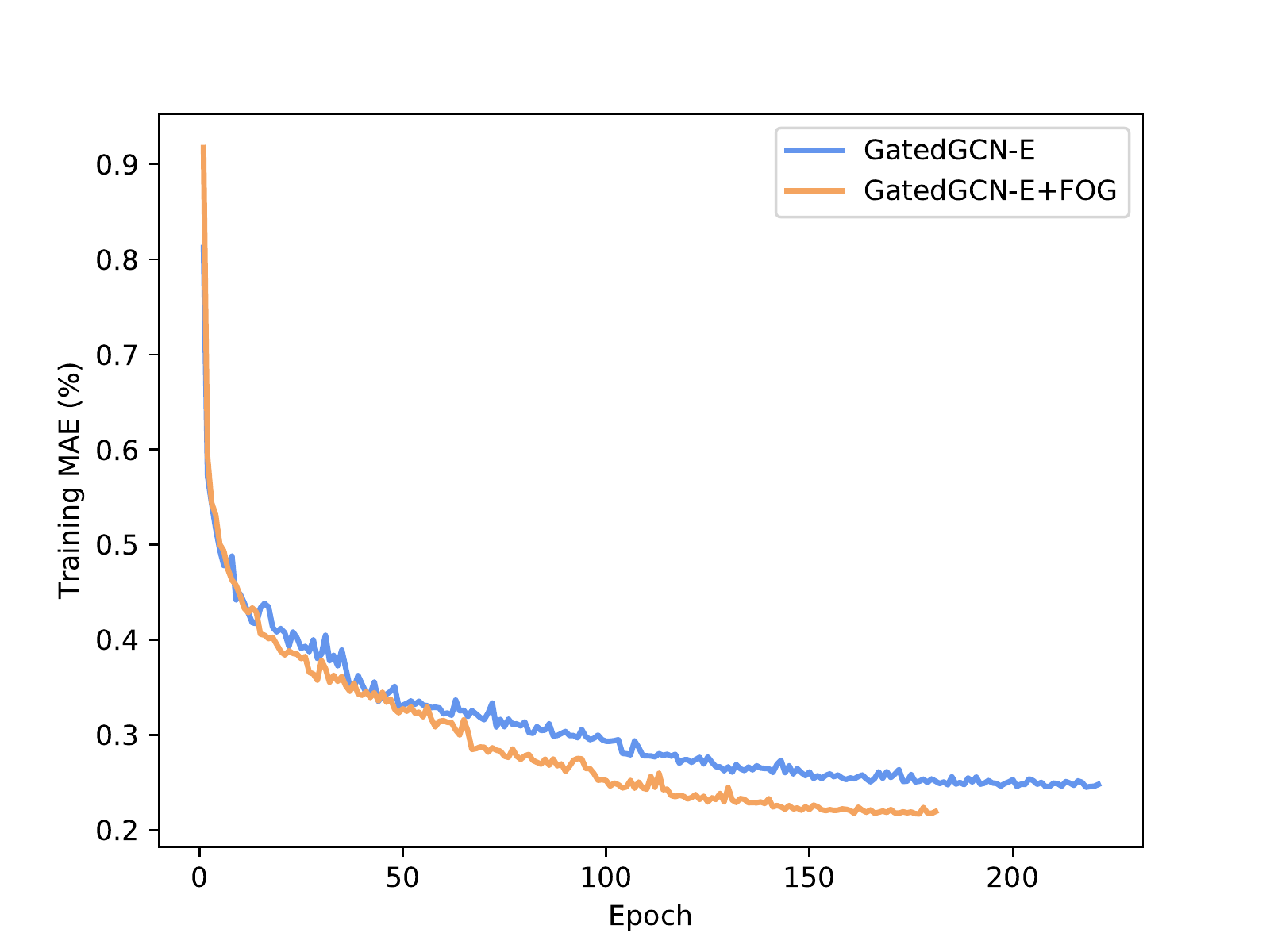}
  \caption{Trainig mean absolute error (MAE) on ZINC with GatedGCN-E and GatedGCN-E+FOG. Both networks have same hyper-parameter settings except hidden layer dimension which is for keeping networks having similar total number of parameters. As the baseline model, GatedGCN-E has a rougher training curve and lower finial performance which will be shown in the experiment section.} \label{fig:GatedGCNEvsGatedEFOG}
\end{figure}

The core part of GNNs is to learn a parametrized network that work as an injection function to project every \textit{multiset} into a single unique representation~\cite{xu2018powerful}. Existing approaches ignore the fact that, given the same multiset, two different center nodes might have significantly different second-order feature spaces while sharing the same first-order feature space. Hence, we propose a novel central node permutation variant aggregation module called \textbf{F}eature c\textbf{O}rrelation \textbf{A}ggregation (FOG) to discover the intrinsic representation ability of a multiset. To address the aforementioned drawbacks of existing GNNs, we use the Kronecker Product of a multiset's feature vector to generate a correlation space (second-order statistical information). In this new space we aggregate the correlation feature vectors as a complementary information based on the aggregation features of existing nodes. Despite two multisets being quite similar, the resulting feature extracted from the 
second-order space is distinctive while the simple first-order aggregation fails to find the difference.  In addition, our FOG may be readily layered on top of any existing GNNs.  As Fig.~\ref{fig:GatedGCNEvsGatedEFOG} shown, FOG helps the GateGCN-E \cite{bresson2017residual} to learn more discriminative feature and achieve lower cross-entropy loss compared to the original networks.


In a nutshell, our contribution in this paper is three-fold.
\begin{enumerate}
    \item We discuss and analyze the weaknesses in current GNN structures, while proposing a new FOG module which dynamically aggregates feature representations from the feature correlation space for each multiset.
    \item Our FOG module is extremely simple and can be inserted in any GNN structure. We carefully design and embed our FOG module into GCNs, GAT, GatedGCNs, GIN, and GraphSAGE with a fewer number of parameters compared to the original models to provide fair comparisons.
    \item  We conduct comprehensive experiments on graph pattern recognition, node classification, graph regression, and edge classification on different types of datasets to validate the effectiveness and compatibility of our module.
\end{enumerate}

\section{Related Works}

\begin{table*}[!ht]
\caption{Commonly used notation in this paper}
\begin{center}
\scalebox{0.8}{
\begin{tabular}{l|l}
\hline
Notation & Descriptions \\
\hline

$f_{base}$ & The function of the base GNN module. \eg, GCN, GAT, GatedGCN, GIN, and GraphSAGE. \\

$\otimes$ & Kronecker product. Given two vectors $\Mat{a} \in \mathbb{R}^{C_\Mat{a}},\Mat{b} \in \mathbb{R}^{C_\Mat{b}}$, the outcome of the Kronecker product is\\
& $\Mat{a} \otimes \Mat{b} = [a_1\Mat{b},a_2\Mat{b},..,a_{C_\Mat{a}}\Mat{b}] \in \mathbb{R}^{C_\Mat{a}C_\Mat{b}}.$ \\
$\relu(\cdot)$ & Rectified linear unit. $\relu(x)=\max(x,0)$. \\

$\concat(\cdot,\cdot)$ & Concatenate two vectors along the channel.\\
$\phi(\cdot)$ & An non-linear activation function.\\
$\bn(\cdot)$ & Batch normalization.\\

\hline
\end{tabular}
}
\end{center}\label{table:ops}
\end{table*}
Our FOG module is used to enhance the representation ability of existing GNN structures, therefore it is closely related to works in that domain.
One of the components in our module related to the way we generate the correlation feature has been investigated in previous studies. In this section, we review methods that are related and discuss their contributions. 

\noindent\textbf{Graph Neural Networks.} GNN was first proposed by Gori~\cite{GNN2005} and have since evolved and been applied in a much wider range of applications. Most of the GNNs are considered from either a spectral or a spatial perspective. For example, by defining the Laplacian matrix for a given graph, the orthogonal graph transform basis can be obtained through applying SVD on the Laplacian matrix. Spectral-based methods are able to filtering the feature vectors using the element-wise Hadamard product in the spatial domain which is equivalent to applying the convolution operation in the graph domain. ChebyNets~\cite{defferrard2016convolutional} approximates the k-th order convolutional kernel by employing Chebyshev polynomials in a recursive way. GCNs~\cite{kipf2016semi} replace the k-th order Chebyshev expansion with a first-order approximation to simplify the structure. 

From the spatial perspective, existing methods focus on the way of picking appropriate neighbors during aggregation. GraphSAGE~\cite{hamilton2017inductive} proposes a sampling algorithm that randomly samples a fixed number of neighbors during training. FastGCN~\cite{chen2018fastgcn} and AdaptGCN~\cite{huang2018adaptive} propose to apply different sampling methods in the local neighborhood to alleviate the exponential growth in the number of locally sampled nodes and thereby speeding up the convergence of the optimization. Instead of dropping nodes, Graph Attention Network (GAT) \cite{velivckovic2017graph} adopts a multi-head attention mechanism \cite{vaswani2017attention} from an NLP task to learn a weight for each central-neighbor pair.

\vspace{2pt}
\noindent\textbf{Feature Correlation Description.} Feature aggregation has been studied extensively for visual tasks, such as image or object retrieval and detection. The Kronecker product is the most widely used operation used to capturing second-order statistical information~\cite{Tuzel2006RegionCovariance,Tuzel2007Human,Lin2015BCNN,NIPS2018_7429,lin2017improved,Shen2018EndtoEndDK, Fang2019BiAttention}. In detection tasks, covariance matrices are investigated to represent the regional descriptors~\cite{Tuzel2006RegionCovariance, Tuzel2007Human}. Lin \etal proposed the bilinear CNN where the outer product of two feature vectors, generated from two non-identical networks, is used to model the local pairwise interactions~\cite{Lin2015BCNN}. A further study in~\cite{lin2017improved} proved that rich correlation statistics could also be obtained by a self outer product. Cross-modality features, \eg, visual and language, can also be fused via a bilinear operation~\cite{NIPS2018_7429}. In~\cite{Fang2019BiAttention}, second-order auto-correlation, generated via a bilinear mapping, is used in an attention block to identify salient regions within the images.

In this paper, the proposed FOG module extracts a more discriminative feature from a correlation space which is constructed by using the Kronecker product between the central node and its neighbours in a graph neural network. This simple, yet effective, module brings significant performance gains across various tasks on graph-structured data over and above existing state-of-the-art GNNs.

\section{Methodology}\label{sec:structure}
\subsection{Notation}
We show a graph $\mathcal{G}$ with nodes $\mathcal{V} = \{v_1, v_2, \ldots, v_n\}$ and edge set $\mathcal{E} \subseteq \mathcal{V} \times \mathcal{V}$ by $\mathcal{G}=(\mathcal{V}, \mathcal{E})$. For each node $v \in \mathcal{V}$, its feature representation is denoted by $\Mat{h}_v \in \mathbb{R}^{C_v}$, where $C_v$ is the channel dimension of the node feature. Given $v$ as a central node, its neighbour set is defined as $\mathcal{N}(v) = \{u_i \in \mathcal{V}: (v, 
u_i) \in \mathcal{E},i=1,2,\cdots, \left | \mathcal{N}(v) \right |\}$. The 
cardinality of the neighbour set of $v$ is shown by 
$\left | \mathcal{N}(v) \right |$. The feature of the edge connecting node $v$ and $u_i$, is denoted by $\Mat{e}_{vu_i} \in \mathbb{R}^{C_e}$, where $C_e$ is the channel dimension of the edge feature. Other notations used in this paper are illustrated in Table \ref{table:ops}.

\begin{figure*}[!ht]
  \centering
  \includegraphics[width=14cm,keepaspectratio]{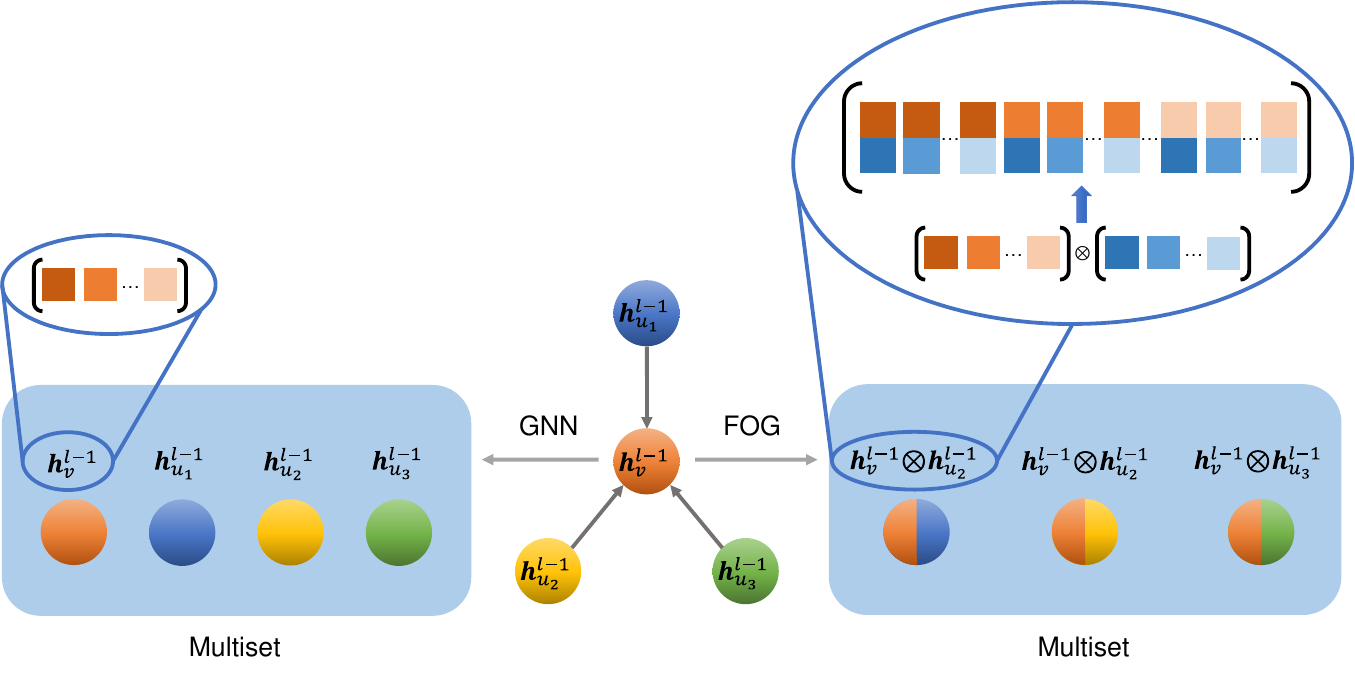}
  \caption{When aggregating neighbors' node features to a central node, GNN only passes the first-order information in the multiset. However, FOG can pass the second-order information through the Kronecker Product operation. Note that, an non-linear layer is applied after the Kronecker Product operation for dimension reduction.} \label{fig:GNNvsFOG}
\end{figure*}

Formally, the $l$-th layer/iteration layer of a GNN can be formulated as,
\begin{align}
\Vec{a}_{v}^l =\text{AGGREGATE}\big(\{\Mat{h}_{u}^{l-1}:u \in \mathcal{N}(v)\}\big)\;,
\end{align}
\begin{align}
\Vec{h}_{v}^l = \text{COMBINE} (\Vec{h}_v^{l-1}, \Vec{a}_{v}^l),
\end{align}
where $\text{AGGREGATE}(\cdot)$ is an aggregation function that aggregating features of neighbors, and $\text{COMBINE}(\cdot)$ means the combination of a central node and its neighbors' aggregated features~\cite{xu2018powerful}. Note that, most GNNs propose a modification of the \text{AGGREGATE} and \text{COMBINE} functions which only obtains the first-order features, e.g. in GAT, \text{AGGREGATE} is a non-linear function, in which neighbors' features are aggregated by weighted summation. Meanwhile, GIN directly uses summation aggregator for neighbors but places a learnable weight on the central node and a non-linear function in the \text{COMBINE}. In contrast, we are taking the feature correlation of the central node and its neighbors into account, and this second-order information can be added on the top of any existing GNN as follow:
\begin{align}
\Vec{h}_{v}^l = \concat\big(\text{COMBINE}(\Vec{h}_v^{l-1}, \Vec{a}_{v}^l), f_{\textrm{FOG}}(\Vec{h}_{v}^{l-1},\{\Vec{h}_{u_i}^{l-1}\})\big),
\label{eq:GNNFOG}
\end{align}
where the $f_{\textrm{FOG}}(\cdot,\cdot)$ is our FOG module and will be described in details in the following section. In Equation \ref{eq:GNNFOG}, our module is working in a different path from the previous methods. In this scenario, FOG generates a more discriminative and diversified feature vector that describes the correlation information between the central node and its neighbors as Figure \ref{fig:GNNvsFOG} shows. This mechanism enables FOG to not only introduce complementary information to existing GNNs architectures and prevent the GNNs from over-smoothing to some extent, but also be a powerful independent architecture.



\subsection{FOG module}

Let  the feature vector of a
node 
$v$ in layer ${l-1}$  be $\Mat{h}_v^{l-1} \in \mathbb{R}^{C_{in}}$. 
The neighbors of $v$, shown by $u_i \in \mathcal{N}(v)$, form a neighbor feature-set $\mathbb{H}_v^{l-1} = \{\Vec{h}_{u_i}^{l-1}\}$, where $\Vec{h}_{u_i}^{l-1} \in \mathbb{R}^{C_{in}}$.
We first perform a non-linear mapping on $\Vec{h}_{v}^{l-1}$ and $\Vec{h}_{u_i}^{l-1}$ to a possibly lower-dimensional space via $\Mat{W}_1^l \in \mathbb{R}^{C_{h1} \times C_{in}}$, and a $\relu$ function. Applied by another mapping $\Mat{W}_2^l \in \mathbb{R}^{C_{h2} \times C_{h1}}$ with $\relu$, the dimensions of neighbor features are further reduced as:

\begin{align}
\tilde{\Vec{h}}_{v}^l = \bn\big(\relu(\Mat{W}_1^l\Mat{h}_{v}^{l-1})\big)\;,
\end{align}
\begin{align}
\tilde{\Vec{h}}_{u_i}^l = \bn\bigg(\relu\Big(\Mat{W}_2^l\bn\big(\relu(\Mat{W}_1^l\Mat{h}_{u_i}^{l-1})\big)\Big)\bigg)\;.
\end{align}

This is followed by computing the Kronecker product between the centre node representation (\ie, $\tilde{\Vec{h}}_{v}^l$)
and the corresponding neighbor nodes (\ie, $\tilde{\Vec{h}}_{u_i}^l$) as follows:

\begin{align}
    \mathbb{R}^{C_{h1}C_{h2}} \ni f_{kp}\big(\tilde{\Vec{h}}_{v}^l,\tilde{\Vec{h}}_{u_i}^l\big)
    = \tilde{\Vec{h}}_{v}^l \otimes \tilde{\Vec{h}}_{u_i}^l\;.
    \label{eqn:KProduct}
\end{align}

In essence, the Kronecker product on $\tilde{\Vec{h}}_{v}^l$ and $\tilde{\Vec{h}}_{u_i}^l$ will enable us to benefit from the correlation information among the center node $v$ and its neighbors. We 
aggregate the correlation among all the neighbors using $f_{\textrm{agg}}(\cdot)$ (\ie, summation) 
into a single vector of dimensionality $C_{h1}C_{h2} $, as: 
\begin{align}
f_{\textrm{agg}}(\{\tilde{\Vec{h}}_{u_i}^l\}) = \sum_{i=1}^N f_{kp}\big(\tilde{\Vec{h}}_{v}^l,\tilde{\Vec{h}}_{u_i}^l\big)\;.
\end{align}
Finally, we pass the aggregated information to a linear mapping $\Mat{W}_{vu}^l \in \mathbb{R}^{C_{p} \times C_{h1}C_{h2}}$ and obtain the output of the FOG module as

\begin{align}
\Mat{p}_{v}^l=f_{FOG}(\Mat{h}_{v}^{l-1},\{\Mat{h}_{u_i}^{l-1}\})=\Mat{W}_{vu}^lf_{\textrm{agg}}(\{\tilde{\Vec{h}}_{u_i}^l\}).
\end{align}
For a plain FOG model, the final output $\Mat{h}_v^l \in \mathbb{R}^{C_{q}}$ is obtained as:

\begin{align}
\Mat{h}_{v}^l= \relu(\Mat{p}_{v}^l).
\end{align}

The FOG has the ability to improve a base GNN module by adding additional second-order information once combined with it. Given a base GNN module with functionality $f_{base}^l$, we can also attain a representation for node $v$ as
\begin{align}
 \mathbb{R}^{C_{q}} \ni \Mat{q}_{v}^l=f_{base}^l(\Mat{h}_{v}^{l-1},\{\Mat{h}_{u_i}^{l-1}\}).
\end{align}

After concatenating $\Mat{p}_{v}^l$ and $\Mat{q}_{v}^l$ along the channel, and subsequently passing the result to an activation function $\phi(\cdot)$ that is used in the original base GNN, the final output $\Mat{h}_v^l \in \mathbb{R}^{C_{out}}$ is obtained as:

\begin{align}
\Mat{h}_{v}^l= \phi\big(\concat(\Mat{p}_{v}^l,\Mat{q}_{v}^l)\big),
\label{eq:hv}
\end{align}
where $C_{out}=C_{p}+C_{q}$.

As Equation \ref{eq:hv} shows, with a minimal amount of changes by concatenating $\Mat{p}_{v}^l$ and $\Mat{q}_{v}^l$, the FOG module can be integrated into most existing GNNs thereby introducing correlation information boosting the performance. In this paper, we introduced the FOG module into five state-of-the-art modules including GCN, GAT, GatedGCN, GIN, and GraphSAGE. Furthermore, we also evaluate FOG with a 2-layer GNN as proposed by \cite{garcia2018fewshot} on a few-shot learning task.

\section{Experiments}

In this section, most experiments are conducted across various datasets based on the benchmarking procedure proposed by \cite{dwivedi2020benchmarking}. This benchmark covers most applications of GNNs, \ie, graph pattern recognition, semi-supervised graph clustering, graph regression, and edge classification. The five aforementioned GNN baselines reported by Dwivedi \etal are compared with their FOG counterparts, along with a FOG-only model, under similar training protocols as in \cite{dwivedi2020benchmarking}. By introducing FOG to a GNN based few-shot learning problem proposed by Garcia and Bruna \cite{garcia2018fewshot}, we explore the possibility of applying FOG on a computer vision task. Furthermore, three more experiments are conducted in ablation study and further discussion to demonstrate the parameter-reducing potential, and limitation of FOG. 

\subsection{Datasets}

\noindent{\textbf{SBM PATTERN and CLUSTER.}} We use the SBM datasets, produced by the stochastic block model, for graph pattern recognition and semi-supervised graph clustering tasks. These two datasets are proposed by \cite{dwivedi2020benchmarking}. The goal of the SBM PATTERN set is to identify a fixed graph pattern embedded in a large graph. This set comprises 10K items in the train set, 2K in the validation set and 2K in the test set. Each graph has on average 117.47 nodes and 2 classes. The SBM CLUSTER is used to evaluate the performance of the model on a semi-supervised clustering task. Given one prior known node label per class, the model learns to gather nodes that belong to the same class according to their connectivity. The number of graphs are 10K for train, 1K for validation, and 1K for test in this dataset. The average number of nodes in each graph is 117.20 and the number of classes is 6.    

\noindent{\textbf{ZINC.}} The ZINC dataset is used to regress the molecular constrained solubility~\cite{ZINC_data}. The number of train, validation and test graphs are 10K, 1K and 1K respectively, with each graph containing 23.16 nodes on average. For each graph, the node features are the type of atoms and edge features are type of bonds.

\noindent{\textbf{TSP.}} We employ the TSP dataset generated by \cite{dwivedi2020benchmarking} to evaluate the edge classification performance of our method. The ability of the algorithm to solving NP-hard combinatorial optimization problems are evaluated on this dataset. The node feature represents the coordinates of a node in a unit square. The train, validation and test sets are split into 10K, 1K and 1K graphs, respectively. Each graph has a different number of nodes. The average number is 275.76.

\noindent{\textbf{MiniImageNet.}}
In our experiments, miniImageNet proposed by \cite{miniImageNet} is used for a few-shot learning task. This dataset contains 100 classes and 600 images for each class. Following \cite{miniImageNet_split}, we use a 64/16/20 split for training/validation/testing on 5-way 1-shot and 5-way 5-shot tasks.

\noindent{\textbf{IMDB-MULTI.}}
IMDB-MULTI is a graph classification benchmark proposed by Yanardag and Vishwanathan \cite{DeepGraphKernels}. In this social network dataset, each actor/actress' ego-network forms a graph that represents one of three genres. Actors/actress are represented as nodes and connected if they performed in the same movie. There are 1500 graphs in this dataset.

\subsection{Implementation and Evaluation}  \label{Implementation}

To verify the superior performance of our proposed FOG module, we evaluate FOG against various baselines mentioned in \textsection~\ref{sec:structure}. Empirical results reveal that the FOG module brings a significant performance gain over the baseline GNNs, with a fewer number of parameters. This observation suggests that the FOG module indeed benefits from the feature generated by the correlation feature space and non-linearity.

For each dataset, the GNN baseline and its FOG counterpart share the same training protocol. To thoroughly verify the effectiveness of our proposed method, if not otherwise mentioned, we used a grid search to find the optimized learning rate in $\{1e^{-2},5e^{-3},1e^{-3},5e^{-4}\}$ and weight decay in $\{1e^{-3},1e^{-6},0\}$, and evaluated the network 10 times using different random seeds, more than that in \cite{dwivedi2020benchmarking}. The Adam optimizer is used in all tasks with $\beta_1=0.9$, and $\beta_2=0.999$. In order to verify that the performance gain indeed comes from our proposed algorithm, the number of parameters of the FOG-only model and FOG-equipped GNN is similar to the baseline GNN. All GNNs evaluated on \cite{dwivedi2020benchmarking} use residual connections \cite{He2016ResNet}, and batch normalization \cite{Ioffe2015BN}. We omit the self loop in the input graphs across all experiments. We use the PyTorch~\cite{Pytorch} deep learning package to implement our algorithm based on the code provided by \cite{dwivedi2020benchmarking}. All experiments are trained on an Nvidia Tesla V100 16GB GPU. By default, 2 layers of a GNN module are used on miniImageNet and 4 layers are used on the other datasets. The detailed architectures and hyper-parameter settings are provided in the supplementary material.  

\vspace{2pt}
\noindent{\textbf{SBM PATTERN and CLUSTER.}} The patience value is set to 5. After 5 epochs with no improvement of the loss, the learning rate will be reduced by a factor of 2. The training progress will stop when the learning rate is smaller than $1e^{-5}$.  All final node features are passed to a 3-layer multilayer perceptron (MLP) including a classification layer that uses the cross-entropy loss to obtain a prediction for each node. The performance metric is the average accuracy between predicted and ground-truth labels.

The evaluation on the SBM dataset is shown in Table~\ref{table:SBM_PATTERN} and \ref{table:SBM_CLUSTER}. In the graph pattern recognition task, the model with only the FOG module surpasses all architectures by a maximum of 21.783\% in terms of accuracy when comparing to GCN (see Table~\ref{table:SBM_PATTERN}). Introducing FOG to GCN, GAT, GatedGCN, and GraphSAGE provide boost to the original structures. However, a performance drop is observed when combining FOG with GIN. We conjecture that the way GIN adds the learnable weights in the COMBINE part conflicts with our FOG.

In the semi-supervised graph clustering task, our FOG module also shows its potential to bring performance gains across different baselines (see Table~\ref{table:SBM_CLUSTER}). Furthermore, plugging in the FOG module consistently improves the accuracy across all baselines and pushes all of them even higher than the FOG-only model and corresponding baseline model. The FOG module improves the accuracy of GCN, GAT, GatedGCN, GIN, and GraphSAGE by 7.334\%, 3.089\%, 4.079\%, 3.624\%, and 9.689\% respectively.

\begin{table}[!ht]
\caption{Performance on the SBM PATTERN test set (higher is better). \textbf{Bold}: the best model between the baseline GNN and the corresponding FOG-equipped GNN. \textbf{\textcolor{red}{Red}}: the best model overall.}
\begin{center}
\scalebox{1}{
\begin{threeparttable}
\begin{tabular}{l|c|c}
\hline
Model & \#Param & Acc(\%)$\pm$s.d.$\uparrow$ \\ 
\hline
FOG & 99,046 & \textbf{\textcolor{red}{85.663 $\pm$ 0.025}} \\ 

\hline
GCN & 100,923 & 63.880 $\pm$ 0.074 \\  
GCN+FOG & 101,026 & \textbf{{85.663 $\pm$ 0.027}} \\

\hline
GAT & 109,936 & 75.824 $\pm$ 1.823 \\  
GAT+FOG & 101,346 & \textbf{85.654 $\pm$ 0.023} \\

\hline
GatedGCN & 104,003 & 84.480 $\pm$ 0.122 \\
GatedGCN+FOG & 102,050 & \textbf{{85.454 $\pm$ 0.090}} \\

\hline
GIN & 100,884 & \textbf{85.590 $\pm$ 0.011} \\
GIN+FOG & 99,234 & 85.524 $\pm$ 0.029 \\

\hline
GraphSAGE & 101,739 & 50.516 $\pm$ 0.001 \\
GraphSAGE+FOG & 95,679 & \textbf{85.578 $\pm$ 0.061} \\

\hline
\end{tabular}
\end{threeparttable}
}
\end{center}\label{table:SBM_PATTERN}
\end{table}

\begin{table}[!ht]
\caption{Performance on the SBM CLUSTER test set (higher is better). \textbf{Bold}: the best model between the baseline GNN and the corresponding FOG-equipped GNN. \textbf{\textcolor{red}{Red}}: the best model overall.}
\begin{center}
\scalebox{1}{
\begin{threeparttable}
\begin{tabular}{l|c|c}
\hline
Model & \#Param & Acc(\%)$\pm$s.d. $\uparrow$ \\ 
\hline
FOG & 99,770 & 58.655 $\pm$ 0.974 \\ 

\hline
GCN & 101,655 & 53.445 $\pm$ 2.029 \\  
GCN+FOG & 101,830 & \textbf{60.779$ \pm$ 0.262} \\
\hline
GAT & 110,700 & 57.732 $\pm$ 0.323 \\  
GAT+FOG & 102,150 & \textbf{{60.821 $\pm$ 1.106}} \\

\hline
GatedGCN & 104,355 & 60.404 $\pm$ 0.419 \\
GatedGCN+FOG &  102,374 & \textbf{\textcolor{red}{64.483 $\pm$ 0.331}} \\

\hline
GIN & 103,544 & 58.384 $\pm$ 0.236 \\
GIN+FOG &  102,806 & \textbf{62.008 $\pm$ 0.591} \\

\hline
GraphSAGE & 102,187 & 50.454 $\pm$ 0.145 \\
GraphSAGE+FOG & 96,171 & \textbf{60.143 $\pm$ 0.414} \\
\hline
\end{tabular}
\end{threeparttable}
}
\end{center}\label{table:SBM_CLUSTER}
\end{table}

\noindent{\textbf{ZINC.}} The same training protocol as for SBM is used here, and the patience value is set to 10. Instead of classification, in ZINC we are targeting a regression task, where a 3-layer MLP is followed after the GNN's last layer to approximate the ground truth. The mean absolute error (MAE) between the predicted and the ground-truth constrained solubility is applied as a loss function, as well as the performance metric.

In the graph regression task, our FOG module improves over most baseline GNNs with respect to the MAE value by a large margin, shown in Table~\ref{table:zinc}. In particular, the GCN+FOG model improves 33.116\% over GCN and exceeds all baselines. Considering that the FOG-only model notably has a lower performance than most FOG-equipped baseline GNNs, the FOG module can provide more complementary information to baseline GNNs on this dataset. However, introducing FOG to GIN suffers a performance drop, which is similar to SBM PATTERN as mentioned above.

\vspace{2pt}
\noindent{\textbf{TSP.}} The following changes are made when adapting the SBM training protocol: (a) The patience value is set to 10. (b) The F1 score only considers the positive class for the performance metric. (c) Edge features are passed to a 3-layer MLP where the last layer is a classification layer. Cross-entropy loss is used for this edge classification task.

The edge classification task also benefits from our FOG module. As shown in Table~\ref{table:tsp}, our FOG module consistently improves over the baseline GNNs. For example, the GCN+FOG improves 9.206\% F1 over GCN. In addition, the best performance architecture, GatedGCN+FOG, outperforms the state-of-the-art baseline (\ie, GatedGCN) by 2.351\% F1 score when using edge features. 

\vspace{2pt}
\noindent{\textbf{MiniImageNet.}} The experiments on this dataset are based on the networks proposed by \cite{garcia2018fewshot}. One query image and its support set are formed as a fully-connected graph where each node is an image. Each architecture has two parts: (1) a 5-layer CNN generates a representation vector as a node feature. (2) a 2-layer GNN passes representations among nodes to predict the label of the query image among the images in the support set. We only concatenate FOG to the first GNN layer for the FOG-equipped version and replace the first GNN layer with FOG for the FOG-only version. All architectures in this task are trained four times with cross-entropy loss and four random seeds in an end-to-end fashion.

The evaluation results of 5-way 1-shot and 5-way 5-shot tasks are shown in Table~\ref{table:miniImageNet}. The original GNN can still gain higher performance after concatenating with FOG.

\begin{table}[!ht]
\caption{Performance on the ZINC test set (lower is better). ~\textbf{Bold}: the best model between the baseline GNN and the corresponding FOG-equipped GNN. \textbf{\textcolor{red}{Red}}: the best model overall.}
\begin{center}
\scalebox{1}{
\begin{threeparttable}
\begin{tabular}{l|c|c}
\hline
Model & \#Param & MAE$\pm$s.d.$\downarrow$ \\ 
\hline
FOG & 101,668 & 0.362 $\pm$ 0.016 \\

\hline
GCN & 103,077 & 0.459 $\pm$ 0.006 \\
GCN+FOG & 102,809 & \textbf{0.307 $\pm$ 0.013}  \\

\hline
GAT & 102,385 & 0.475 $\pm$ 0.007 \\
GAT+FOG & 105,305 & {\textbf{0.333 $\pm$ 0.016}} \\

\hline
GatedGCN & 105,735 & 0.435 $\pm$ 0.011 \\
GatedGCN+FOG & 103,633 & \textbf{0.318 $\pm$ 0.022} \\

\hline
GatedGCN-E\tnote{*} & 105,875 & 0.375 $\pm$ 0.003 \\
GatedGCN-E+FOG\tnote{*} & 103,761 &
\textcolor{red}{\textbf{0.271 $\pm$ 0.030}} \\

\hline
GIN & 103,079 & \textbf{0.387 $\pm$ 0.015} \\
GIN+FOG & 102,189 & 0.401 $\pm $0.020  \\

\hline
GraphSAGE & 94,977 & 0.468 $\pm$ 0.003 \\
GraphSAGE+FOG & 94,477 & \textbf{0.317 $\pm$ 0.011} \\
\hline
\end{tabular}
\begin{tablenotes}
\item[*] -E stands for using the molecule bond type as edge feature.
\end{tablenotes}
\end{threeparttable}
}
\end{center}
\label{table:zinc}
\end{table}

\begin{table}[!ht]
\caption{Performance on the TSP test set (higher is better). \textbf{Bold}: the best model between the baseline GNN and the corresponding FOG-equipped GNN. \textbf{\textcolor{red}{Red}}: the best model overall.}
\begin{center}
\scalebox{1}{
\begin{threeparttable}
\begin{tabular}{l|c|c}
\hline
Model & \#Param & F1$\pm$s.d. $\uparrow$\\ 
\hline
FOG & 96,386 & 0.677 $\pm$ 0.004 \\

\hline
GCN & 95,702 & 0.630 $\pm$ 0.001 \\
GCN+FOG & 93,465 & \textbf{0.688 $\pm$ 0.007} \\

\hline
GAT & 96,182 & 0.671 $\pm$ 0.002 \\ 
GAT+FOG & 96,350 & \textbf{0.713 $\pm$ 0.002} \\

\hline
GatedGCN & 97,858 & 0.791 $\pm$ 0.003 \\ 
GatedGCN+FOG & 95,456 & \textbf{0.822 $\pm$ 0.002} \\

\hline
GatedGCN-E\tnote{*} & 97,858 & 0.808 $\pm$ 0.003 \\ 
GatedGCN-E+FOG\tnote{*} & 95,456 & \textbf{\textcolor{red}{0.827 $\pm$ 0.002}} \\

\hline
GIN & 99,002 & 0.656 $\pm$ 0.003 \\
GIN+FOG & 94,046 & \textbf{0.706 $\pm$ 0.010} \\

\hline
GraphSAGE & 99,263 & 0.665 $\pm$ 0.003 \\
GraphSAGE+FOG & 97,007 & \textbf{0.689 $\pm$ 0.002} \\

\hline
\end{tabular}
\begin{tablenotes}
\item[*] -E stands for using the pairwise distance as an edge feature.
\end{tablenotes}
\end{threeparttable}
}
\end{center}\label{table:tsp}
\end{table}

\begin{table}[!ht]
\caption{Performance on the miniImageNet test set with 95\% confidence intervals (higher is better). \textbf{\textcolor{red}{Red}}: the best model overall. \#Param: the number of parameters in GNN parts}
\begin{center}
\scalebox{0.98}{
\begin{threeparttable}
\begin{tabular}{l|c|c|c}
\hline
\multirow{2}{*}{Model} & \multirow{2}{*}{\#Param} & 5-way 1-shot & 5-way 5-shot\\
\cline{3-4}
& &Acc (\%) $\uparrow$&Acc (\%) $\uparrow$\\
\hline
FOG & 312,282 & 50.70 $\pm$ 0.69  & 65.64 $\pm$ 0.59 \\  
\hline
GNN & 335,994 & 50.33 $\pm$ 0.36  & 66.41 $\pm$ 0.63 \\  
GNN+FOG & 323,760 & \textbf{\textcolor{red}{50.71 $\pm$ 0.55}}  & \textbf{\textcolor{red}{66.77 $\pm$ 0.49}} \\
\hline
\end{tabular}
\end{threeparttable}
}
\end{center}\label{table:miniImageNet}
\end{table}

\subsection{Ablation Study}
\noindent{\textbf{Dimensionality of hidden layers.}} According to the notable boost when introducing FOG into the state-of-the-art module, it demonstrates the ability of FOG reducing the total number of parameters of the original architectures while still maintaining the performance. To demonstrate this point, we choose the simplest architecture: GCN, and the best performing architecture: GatedGCN. Their FOG-equipped counterparts with different number of parameters are evaluated on ZINC by using the same training protocol as in \textsection~\ref{Implementation}. All models use the same learning rate and weight decay found via previous experiments. Their number of parameters are roughly divided from 25\% to 100\% by adjusting hidden layer dimensions according to their baseline models.

As Table \ref{table:DimStudy} shows, all variations are better than their corresponding baseline models. For the best performing baseline on ZINC, FOG boosts the performance by 11.733\% compared to GatedGCN by only using about 25\% of the parameters. Both 25\% architectures are still better than the other baseline models in Table \ref{table:zinc}.

\begin{table}[!ht]
\caption{Performance of GCN, GatedGCN, and their corresponding FOG-equipped variants with different number of hidden layer dimensions on the ZINC test set (lower is better).}
\begin{center}
\scalebox{0.85}{
\begin{threeparttable}
\begin{tabular}{l|c|c|c}
\hline
Model & \#Param & \#Param ratio & MAE$\pm$s.d. $\downarrow$\\ 
\hline
GCN & 103,077 & 100\% & 0.459 $\pm$ 0.006 \\
\hline
\multirow{4}{*}{GCN+FOG}
 & 102,809 & 100\% & 0.307 $\pm$ 0.013  \\
 & 77,278 & 75\% & {0.322 $\pm$ 0.015}  \\
 & 50,547 & 50\% & {0.337 $\pm$ 0.021}  \\
 & 25,847 & 25\% & {0.352 $\pm$ 0.025}  \\
\hline
GatedGCN-E\tnote{*} & 105,875 & 100\% & 0.375 $\pm$ 0.003 \\
\hline

\multirow{4}{*}{\shortstack[l]{GatedGCN-E \\ +FOG\tnote{*}}}
 & 103,761 & 100\% & 0.271 $\pm$ 0.030 \\
 & 79,165 & 75\% & 0.287 $\pm$ 0.029  \\
 & 49,835 & 50\% & 0.307 $\pm$ 0.016  \\
 & 26,909 & 25\% & 0.331 $\pm$ 0.031  \\
\hline
\end{tabular}
\begin{tablenotes}
\item[*] -E stands for using the pairwise distance as edge feature.
\end{tablenotes}
\end{threeparttable}
}
\end{center}\label{table:DimStudy}
\end{table}

\subsection{Further Discussion}

After demonstrating the strengths of our FOG, we address some limitations here. As discussed above, FOG enjoys the benefits of the node's feature, meanwhile, it is also constrained when the node's feature is absent or uninformative. Therefore, we do further study on the behavior of our FOG on those datasets.

Other techniques, in comparison to ours, focus on graph structural information while ignoring the correlation on feature space. Among them, GIN and GatedGCN are the best architectures on ZINC regardless of using edge feature or not. As a result, we assess both of them, as well as their FOG-equipped versions, alongside a FOG-only design.

Following the setting of \cite{xu2018powerful}, we set all node features to the uninformative value 1 on IMDB-MULTI dataset, where 10-fold cross-validation and only one random seed is applied. Same patience value, learning rate reduction factor, and minimum learning rate are set as experiments on SBM dataset, and a 3-layer MLP is used to predict labels by using final node features. To show the importance of the nodes' feature, we replace the nodes' feature vector on ZINC dataset with single value 1. The other experimental setting is the same as Table~\ref{table:zinc}.


Table \ref{table:Limitation} illustrates that comparing to state-of-the-art models, the notable performance drop when only applying FOG on graphs without node feature. However, GatedGCN+FOG achieves better results comparing to the baseline on both two datasets. This implies FOG can still provide complementary information for the base GNNs while independent FOG may struggle on extracting correlation information from graph structure. The consistent performance drop while concatenating FOG with GIN suggests that the learnable parameter $\epsilon$ in GIN may break the correlation between the central node and its neighbors, which makes these two modules are incompatible in most scenarios. Generally, all networks suffer large performance drop when comparing to Table \ref{table:zinc}. This suggests original node features and edge features have important contribution to this kind of tasks.

\begin{table}[!ht] 
\caption{Performance on the ZINC without node feature test set (lower is better) and IMDB-MULTI validation set (higher is better). ~\textbf{Bold}: the best model between the baseline GNN and the corresponding FOG-equipped GNN. \textbf{\textcolor{red}{Red}}: the best model overall.}
\begin{center}
\scalebox{0.76}
{
\begin{tabular}{l|cc|cc}
\hline
\multirow{2}{*}{Model} & \multicolumn{2}{c|}{ZINC w/o} & \multicolumn{2}{c}{IMDB-M} \\
\cline{2-5}
                      & \#Param        & MAE$\pm$s.d.$\downarrow$  & \#Param      & Acc(\%)$\pm$s.d.$\uparrow$         \\
\hline
FOG          & 101,668  & 1.308 $\pm$ 0.017 & 33,017 & 46.067 $\pm$ 4.848    \\
\hline
GatedGCN     & 105,735  & 1.296 $\pm$ 0.024 & 34,663 & 50.933 $\pm$ 4.123    \\
GatedGCN+FOG & 103,633  & \textbf{1.283 $\pm$ 0.014} & 33,303 & \textcolor{red}{\textbf{52.067 $\pm$ 3.508}}    \\
\hline
GIN          & 103,079  & \textcolor{red}{\textbf{1.277 $\pm$ 0.016}} & 35,411 & \textbf{48.400 $\pm$ 4.716}    \\
GIN+FOG      & 102,189  & 1.293 $\pm$ 0.029 & 34,805 & 47.933 $\pm$ 3.521    \\    
\hline
\end{tabular}
}
\end{center}\label{table:Limitation}
\end{table}
\section{Conclusion}

We have presented a new aggregation module, namely FOG, which introduces the feature correlation between the central node and its neighbors into the central node's feature representation. We, firstly, summarize the way of how existing GNNs work and generalize them into the same presentation. Distinguishing from existing methods, our method is compatible with most existing methods so that it can be inserted on top of those GNNs. Furthermore, our extensive experiments on different datasets have shown that FOG is able to help the GNNs to get more discriminative features and achieve better performance in various tasks, e.g. graph pattern recognition, node classification, graph regression, and edge classification. Furthermore, we also provide a thorough analysis of the limitation and applicability of our method. In the future, we will explore other types of function which is sensitive to the change of central node, and study on how it affects the GNNs.

{\small
\bibliographystyle{ieee_fullname}
\bibliography{main.bib}

\begin{thebibliography}{10}\itemsep=-1pt

\bibitem{Pytorch}
Brandon Amos, Ivan Jimenez, Jacob Sacks, Byron Boots, and J.~Zico Kolter.
\newblock Differentiable {MPC} for {End}-to-end {Planning} and {Control}.
\newblock In {\em Advances in Neural Information Processing Systems},
  volume~31, pages 8289--8300, 2018.

\bibitem{bresson2017residual}
Xavier Bresson and Thomas Laurent.
\newblock Residual {Gated} {Graph} {Conv}{Nets}.
\newblock {\em arXiv preprint arXiv:1711.07553}, 2017.

\bibitem{chen2018fastgcn}
Jie Chen, Tengfei Ma, and Cao Xiao.
\newblock Fast{GCN}: {Fast} {Learning} with {Graph} {Convolutional} {Networks}
  via {Importance} {Sampling}.
\newblock In {\em International Conference on Learning Representations}, 2018.

\bibitem{defferrard2016convolutional}
Micha{\"e}l Defferrard, Xavier Bresson, and Pierre Vandergheynst.
\newblock Convolutional {Neural} {Networks} on {Graphs} with {Fast} {Localized}
  {Spectral} {Filtering}.
\newblock In {\em Advances in Neural Information Processing Systems}, 2016.

\bibitem{imagenet_cvpr09}
Jia Deng, Wei Dong, Richard Socher, Li-Jia Li, Kai Li, and Fei-Fei Li.
\newblock {ImageNet: A Large-Scale Hierarchical Image Database}.
\newblock In {\em 2009 IEEE Conference on Computer Vision and Pattern
  Recognition}, 2009.

\bibitem{dwivedi2020benchmarking}
Vijay~Prakash Dwivedi, Chaitanya~K Joshi, Thomas Laurent, Yoshua Bengio, and
  Xavier Bresson.
\newblock Benchmarking {Graph} {Neural} {Networks}.
\newblock {\em arXiv preprint arXiv:2003.00982}, 2020.

\bibitem{Fang2019BiAttention}
Pengfei Fang, Jieming Zhou, Soumava~Kumar Roy, Lars Petersson, and Mehrtash
  Harandi.
\newblock Bilinear {Attention} {Networks} for {Person} {Retrieval}.
\newblock In {\em The IEEE International Conference on Computer Vision}, 2019.

\bibitem{MessagePassing}
Justin Gilmer, Samuel~S. Schoenholz, Patrick~F. Riley, Oriol Vinyals, and
  George~E. Dahl.
\newblock Neural message passing for quantum chemistry.
\newblock In {\em Proceedings of the 34th International Conference on Machine
  Learning}, volume~70 of {\em Proceedings of Machine Learning Research}, pages
  1263--1272, 06--11 Aug 2017.

\bibitem{GNN2005}
M. {Gori}, G. {Monfardini}, and F. {Scarselli}.
\newblock A new model for learning in graph domains.
\newblock In {\em Proceedings in 2005 IEEE International Joint Conference on
  Neural Networks}, 2005.

\bibitem{hamilton2017inductive}
Will Hamilton, Zhitao Ying, and Jure Leskovec.
\newblock Inductive {Representation} {Learning} on {Large} {Graphs}.
\newblock In {\em Advances in Neural Information Processing Systems}, 2017.

\bibitem{He2016ResNet}
K. {He}, X. {Zhang}, S. {Ren}, and J. {Sun}.
\newblock Deep {Residual} {Learning} for {Image} {Recognition}.
\newblock In {\em 2016 IEEE Conference on Computer Vision and Pattern
  Recognition}, 2016.

\bibitem{henaff2015deep}
Mikael Henaff, Joan Bruna, and Yann LeCun.
\newblock Deep {Convolutional} {Networks} on {Graph}-{Structured} {Data}.
\newblock {\em arXiv preprint arXiv:1506.05163}, 2015.

\bibitem{huang2018adaptive}
Wenbing Huang, Tong Zhang, Yu Rong, and Junzhou Huang.
\newblock Adaptive {Sampling} {Towards} {Fast} {Graph} {Representation}
  {Learning}.
\newblock In {\em Advances in neural information processing systems}, 2018.

\bibitem{Ioffe2015BN}
Sergey Ioffe and Christian Szegedy.
\newblock Batch {Normalization}: {Accelerating} {Deep} {Network} {Training} by
  {Reducing} {Internal} {Covariate} {Shift}.
\newblock In {\em Proceedings of the 32nd International Conference on
  International Conference on Machine Learning}, 2015.

\bibitem{NIPS2018_7429}
Jin-Hwa Kim, Jaehyun Jun, and Byoung-Tak Zhang.
\newblock Bilinear {Attention} {Networks}.
\newblock In {\em Advances in Neural Information Processing Systems 31}, 2018.

\bibitem{kipf2016semi}
Thomas~N. Kipf and Max Welling.
\newblock Semi-{Supervised} {Classification} with {Graph} {Convolutional}
  {Networks}.
\newblock In {\em 5th International Conference on Learning Representations,
  {ICLR} 2017}. OpenReview.net, 2017.

\bibitem{leman1968reduction}
AA Leman and B Weisfeiler.
\newblock A reduction of a graph to a canonical form and an algebra arising
  during this reduction.
\newblock {\em Nauchno-Technicheskaya Informatsiya}, 2(9):12--16, 1968.

\bibitem{Lin2015BCNN}
T. {Lin}, A. {RoyChowdhury}, and S. {Maji}.
\newblock Bilinear {CNN} {M}odels for {F}ine-grained {V}isual {R}ecognition.
\newblock In {\em 2015 IEEE International Conference on Computer Vision}, 2015.

\bibitem{lin2014microsoft}
Tsung-Yi Lin, Michael Maire, Serge Belongie, James Hays, Pietro Perona, Deva
  Ramanan, Piotr Doll{\'a}r, and C~Lawrence Zitnick.
\newblock Microsoft {C}{O}{C}{O}: {Common} {Objects} in {Context}.
\newblock In {\em 2014 European conference on computer vision}, 2014.

\bibitem{lin2017improved}
Tsung-Yu Lin and Subhransu Maji.
\newblock {Improved Bilinear Pooling with CNNs}.
\newblock In {\em British Machine Vision Conference}, 2017.

\bibitem{miniImageNet_split}
S. Ravi and H. Larochelle.
\newblock Optimization as a model for few-shot learning.
\newblock In {\em ICLR}, 2017.

\bibitem{ribeiro2017struc2vec}
Leonardo~FR Ribeiro, Pedro~HP Saverese, and Daniel~R Figueiredo.
\newblock struc2vec: Learning {Node} {Representations} from {Structural}
  {Identity}.
\newblock In {\em Proceedings of the 23rd ACM SIGKDD International Conference
  on Knowledge Discovery and Data Mining}, 2017.

\bibitem{garcia2018fewshot}
Victor~Garcia Satorras and Joan~Bruna Estrach.
\newblock Few-shot learning with graph neural networks.
\newblock In {\em International Conference on Learning Representations}, 2018.

\bibitem{Shen2018EndtoEndDK}
Yantao Shen, Tong Xiao, Hongsheng Li, Shuai Yi, and Xiaogang Wang.
\newblock End-to-{E}nd {D}eep {K}ronecker-{P}roduct {M}atching for {P}erson
  {R}e-identification.
\newblock {\em 2018 IEEE/CVF Conference on Computer Vision and Pattern
  Recognition}, 2018.

\bibitem{ZINC_data}
Teague Sterling and John~J. Irwin.
\newblock Zinc15 – {Ligand} {Discovery} for {Everyone}.
\newblock {\em Journal of Chemical Information and Modeling}, 2015.

\bibitem{Tuzel2006RegionCovariance}
Oncel Tuzel, Fatih Porikli, and Peter Meer.
\newblock Region {Covariance}: A {Fast} {Descriptor} for {Detection} and
  {Classification}.
\newblock In {\em Proceedings of the 9th European Conference on Computer
  Vision}, 2006.

\bibitem{Tuzel2007Human}
O. {Tuzel}, F. {Porikli}, and P. {Meer}.
\newblock Human {Detection} via {Classification} on {Riemannian} {Manifolds}.
\newblock In {\em 2007 IEEE Conference on Computer Vision and Pattern
  Recognition}, 2007.

\bibitem{vaswani2017attention}
Ashish Vaswani, Noam Shazeer, Niki Parmar, Jakob Uszkoreit, Llion Jones,
  Aidan~N Gomez, {\L}ukasz Kaiser, and Illia Polosukhin.
\newblock Attention {Is} {All} {You} {Need}.
\newblock In {\em Advances in neural information processing systems}, 2017.

\bibitem{velivckovic2017graph}
Petar Veli{\v{c}}kovi{\'{c}}, Guillem Cucurull, Arantxa Casanova, Adriana
  Romero, Pietro Li{\`{o}}, and Yoshua Bengio.
\newblock {Graph Attention Networks}.
\newblock {\em International Conference on Learning Representations}, 2018.

\bibitem{miniImageNet}
Oriol Vinyals, Charles Blundell, Timothy Lillicrap, koray kavukcuoglu, and Daan
  Wierstra.
\newblock Matching networks for one shot learning.
\newblock In D. Lee, M. Sugiyama, U. Luxburg, I. Guyon, and R. Garnett,
  editors, {\em Advances in Neural Information Processing Systems}, volume~29,
  2016.

\bibitem{xu2018powerful}
Keyulu Xu, Weihua Hu, Jure Leskovec, and Stefanie Jegelka.
\newblock {How Powerful are Graph Neural Networks?}
\newblock In {\em International Conference on Learning Representations}, 2019.

\bibitem{DeepGraphKernels}
Pinar Yanardag and S.V.N. Vishwanathan.
\newblock {Deep Graph Kernels}.
\newblock In {\em Proceedings of the 21th ACM SIGKDD International Conference
  on Knowledge Discovery and Data Mining}, KDD '15, page 1365–1374, 2015.

\end{thebibliography}
}

\begin{appendices}
\section{Module Architecture Details}
\subsection{FOG}
\begin{figure*}[!ht]
  \centering
  \includegraphics[width=16cm,keepaspectratio]{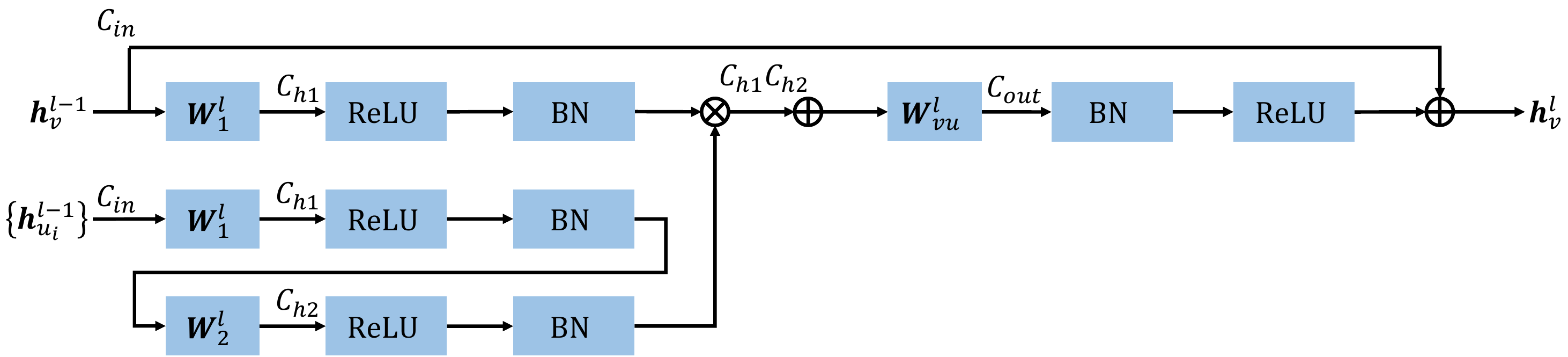}
  \caption{The architecture of FOG module in the $l^{th}$ layer. $C_*$ indicates the dimension of feature vector. $\bn$ is batch normalization.  \textcolor{Black}{$\protect\boldsymbol{\otimes}$} and \textcolor{Black}{$\protect\boldsymbol{\oplus}$} denote the Kronecker product and the element-wise summation functions, respectively..} \label{fig:FOG}
\end{figure*}

In the $l^{th}$ layer of FOG module, the input node feature of the central node $v$ is denoted as $\Mat{h}_v^{l-1}$. The input node feature of the $i^{th}$ neighbor node $u_i$ is denoted as $\Mat{h}_{u_i}^{l-1}$. The architecture of FOG module is illustrated as Figure \ref{fig:FOG}.

\subsection{GCN with FOG} 
As a first-order approximation of ChebyNet \cite{defferrard2016convolutional}, Graph ConvNets (GCN) builds a bridge from spectral based methods to spatial based methods by averaging the features of the first-hop neighbors \cite{kipf2016semi}. 

In the $l^{th}$ layer of GCN module, the input node feature of the central node $v$ is denoted as $\Mat{h}_v^{l-1}$. The input node feature of the $i^{th}$ neighbor node $u_i$ is denoted as $\Mat{h}_{u_i}^{l-1}$. The node  equation of the GCN can be formulated as~\cite{dwivedi2020benchmarking}:

\begin{equation}
{\Mat{h}}_{v}^l=\relu(\sum_{i=1}^{N}{\frac{1}{c_{u_iv}}\Mat{U}^l\Mat{h}_{u_i}^{l-1}}),
\label{con:GCN}
\end{equation}
where $\Mat{U}^l \in \mathbb{R}^{C_{out} \times C_{in}}$, $N=\left | \mathcal{N}(v) \right |$, and $c_{u_iv}=\sqrt{\left | \mathcal{N}(u_i) \right |}
\sqrt{\left | \mathcal{N}(v) \right |}$. 

\begin{figure*}[!ht]
  \centering
  \includegraphics[width=12cm,keepaspectratio]{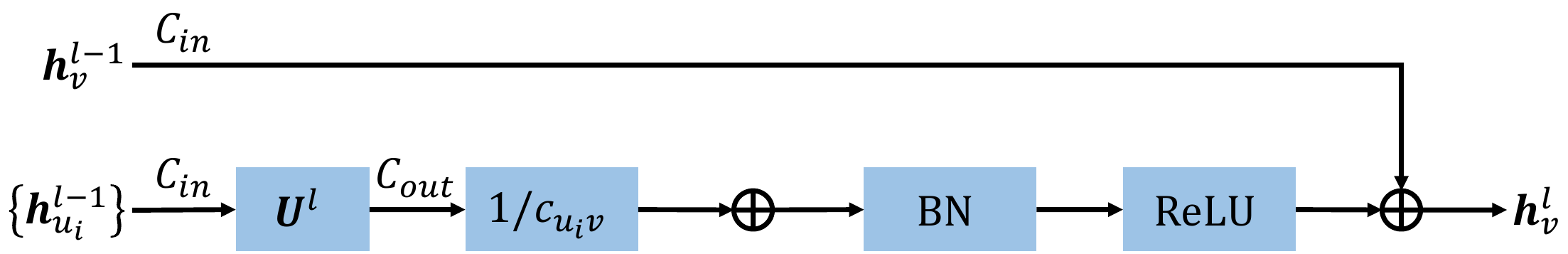}
  \caption{The architecture of GCN module in the $l^{th}$ layer. $C_*$ indicates the dimension of feature vector. $\bn$ is batch normalization.  \textcolor{Black}{$\protect\boldsymbol{\oplus}$} denotes the element-wise summation functions.} \label{fig:GCN}
\end{figure*}

\begin{figure*}[!ht]
  \centering
  \includegraphics[width=15cm,keepaspectratio]{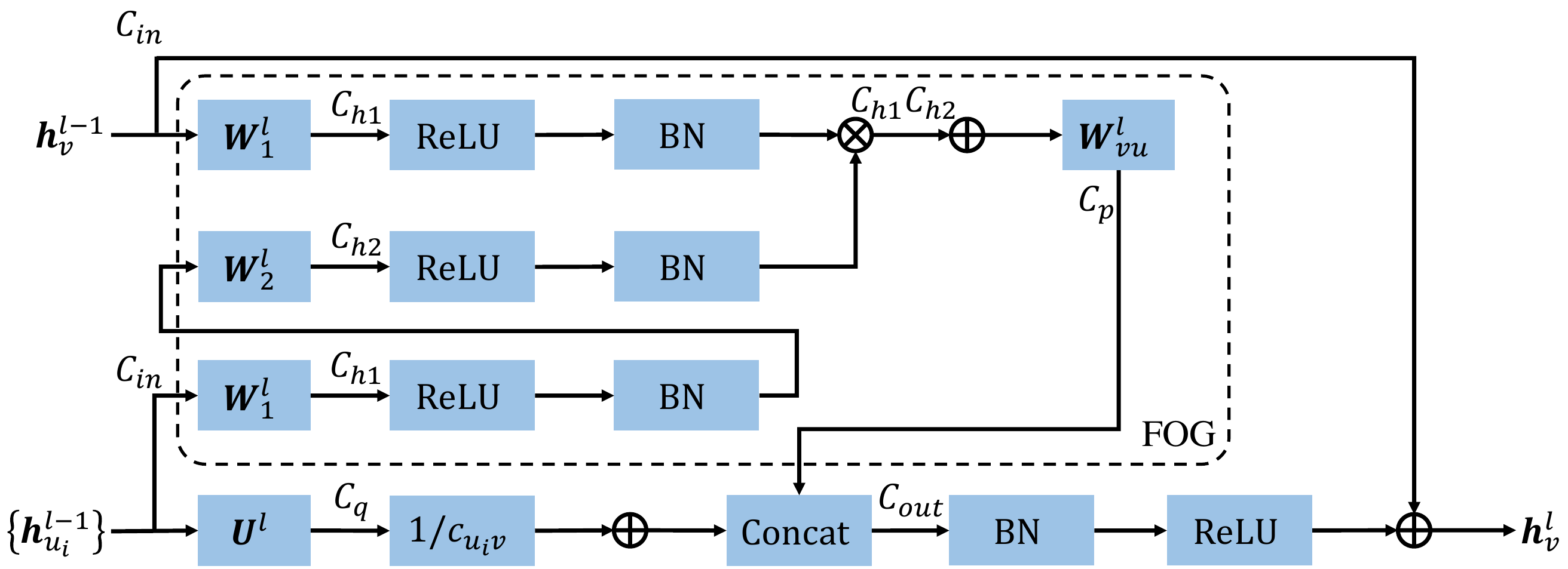}
  \caption{The architecture of GCN+FOG module in the $l^{th}$ layer. $C_*$ indicates the dimension of feature vector. $\concat$ is concatenation along the channel. $\bn$ is batch normalization.  \textcolor{Black}{$\protect\boldsymbol{\otimes}$} and \textcolor{Black}{$\protect\boldsymbol{\oplus}$} denote the Kronecker product and the element-wise summation functions, respectively.} \label{fig:GCN_FOG}
\end{figure*}

After introducing the FOG module into GCN, the functionality of the hybrid GCN with FOG can be written:
\begin{equation}
{\Mat{h}}_{v}^l= \relu\big(\concat(\Mat{p}_{v}^l,\Mat{q}_{v}^l)\big),
\label{con:Bi_GCN}
\end{equation}

\begin{equation}
\Mat{p}_{v}^l=f_{FOG}(\Mat{h}_{v}^{l-1},\{\Mat{h}_{u_i}^{l-1}\}), 
\end{equation}

\begin{equation}
\Mat{q}_{v}^l=\sum_{i=1}^{N}{\frac{1}{c_{u_iv}}\Mat{U}^l\Mat{h}_{u_i}^{l-1}},
\end{equation}
where $\Mat{U}^l \in \mathbb{R}^{{C_q} \times C_{in}}$. The architecture of GCN and GCN+FOG module are illustrated as Figure \ref{fig:GCN} and Figure \ref{fig:GCN_FOG}, respectively.

\subsection{GAT with FOG} 
The Graph Attention Network (GAT) takes a further step towards adaptively aggregating the first-hop neighbors by using a multi-headed attention mechanism \cite{velivckovic2017graph}.

In the $l^{th}$ layer of $K$ heads GAT module, the input node feature of the central node $v$ is denoted as $\Mat{h}_v^{l-1}$. The input node feature of the $i^{th}$ neighbor node $u_i$ is denoted as $\Mat{h}_{u_i}^{l-1}$. The attention coefficients $\alpha_{{v}{u_i}}^{k,l}$ are obtained via:

\begin{equation}
\hat{\alpha}_{{v}{u_i}}^{k,l}=\leakyrelu\big(\Mat{V}^{k,l}\concat(\Mat{U}^{k,l}\Mat{h}_{v}^{l-1},\Mat{U}^{k,l}\Mat{h}_{u_i}^{l-1})\big),
\end{equation}
\begin{equation}
\alpha_{{v}{u_i}}^{k,l}=\softmax_{u_i}({\hat{\alpha}_{{v}{u_i}}^{k,l}})=\frac{\mathrm{exp}(\hat{\alpha}_{{v}{u_i}}^{k,l})}{\sum_{j=1}^{N}{\mathrm{exp}(\hat{\alpha}_{{v}{u_{j}}}^{k,l})}},
\end{equation}
where $k=1,2,\ldots,K$, and $K$ is the number of heads, $\Mat{U}^{k,l} \in \mathbb{R}^{{C_{out}}/{K} \times C_{in}}$, $\Mat{V}^{k,l} \in \mathbb{R}^{{2C_{out}}/{K}}$. 
After the mapping $\Mat{U}^{k,l}$, feature vectors of neighbors are weighted by the corresponding $\alpha_{{v}{u_i}}^{k,l}$ and summed, which is formulated as:

\begin{equation}
\Mat{h}_{v}^l={\overset{K}{\underset{k=1}{\Big|\Big|}}} {\elu(\sum_{i=1}^{N}{\alpha_{vu_i}^{k,l}\Mat{U}^{k,l}\Mat{h}_{u_i}^{l-1}})},
\label{con:GAT}
\end{equation}
where $N=\left | \mathcal{N}(v) \right |$, $||_{k=1}^K$ represents the concatenation of $K$ heads along the channel.

\begin{figure*}[!ht]
  \centering
  \includegraphics[width=15cm,keepaspectratio]{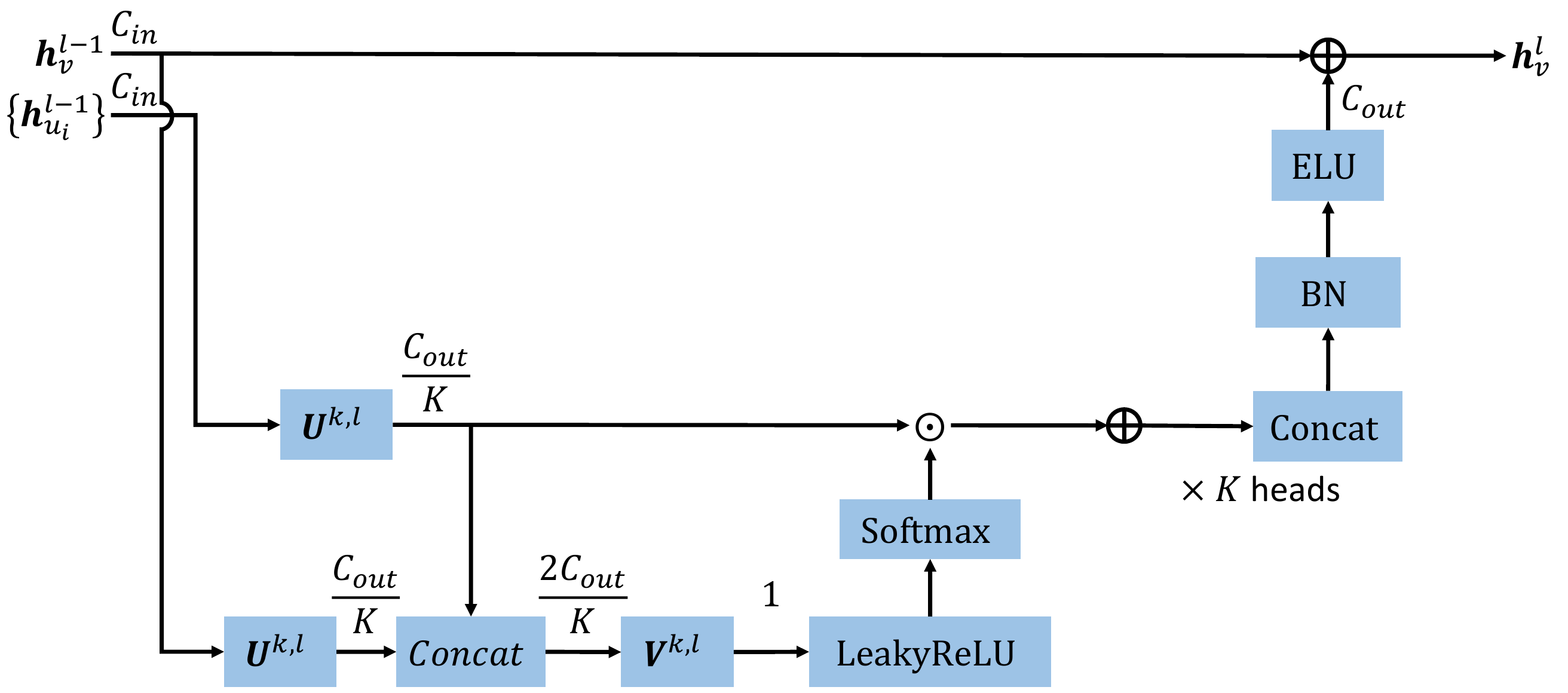}
  \caption{The architecture of GAT module in the $l^{th}$ layer. $C_*$ indicates the dimension of feature vector. $\concat$ is concatenation along the channel. $\bn$ is batch normalization.  \textcolor{Black}{$\protect\boldsymbol{\odot}$} denotes the Hadamard product. \textcolor{Black}{$\protect\boldsymbol{\oplus}$} denotes the element-wise summation functions.} \label{fig:GAT}
\end{figure*}

\begin{figure*}[!ht]
  \centering
  \includegraphics[width=\textwidth,keepaspectratio]{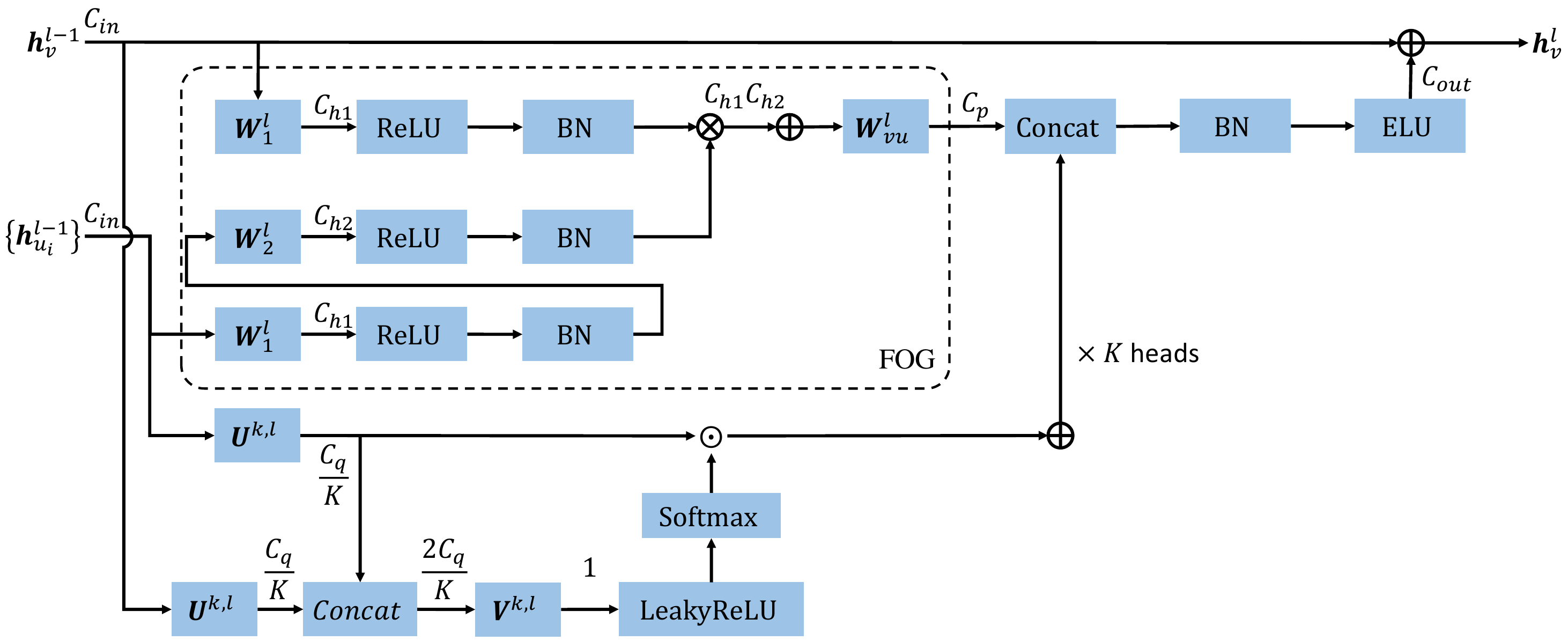}
  \caption{The architecture of GAT+FOG module in the $l^{th}$ layer. $C_*$ indicates the dimension of feature vector. $\concat$ is concatenation along the channel. $\bn$ is batch normalization.  \textcolor{Black}{$\protect\boldsymbol{\otimes}$} and \textcolor{Black}{$\protect\boldsymbol{\oplus}$} denote the Kronecker product and the element-wise summation functions, respectively. \textcolor{Black}{$\protect\boldsymbol{\odot}$} denotes the Hadamard product.} \label{fig:GAT_FOG}
\end{figure*}

When concatenating with the FOG module, the dimensions of $\Mat{U}^{k,l}$ and $\Mat{V}^{k,l}$ are changed to $\mathbb{R}^{{C_q}/{K} \times C_{in}}$ and $\mathbb{R}^{{2C_q}/{K}}$, respectively. Meanwhile, Equation \eqref{con:GAT} is changed to:

\begin{equation}
\Mat{h}_{v}^l=\elu\big(\concat(\Mat{p}_{v}^l,\Mat{q}_{v}^l)\big),
\label{con:Bi_GAT}
\end{equation}
\begin{equation}
\Mat{p}_{v}^l=f_{FOG}(\Mat{h}_{v}^{l-1},\{\Mat{h}_{u_i}^{l-1}\}),
\end{equation}
\begin{equation}
\Mat{q}_{v}^l={\overset{K}{\underset{k=1}{\Big|\Big|}}}{(\sum_{i=1}^{N}{\alpha_{vu_i}^{k,l}\Mat{U}^{k,l}\Mat{h}_{u_i}^{l-1}})}.
\end{equation}
The architecture of GAT and GAT+FOG module are illustrated as Figure \ref{fig:GAT} and Figure \ref{fig:GAT_FOG}, respectively.

\subsection{GatedGCN with FOG} 
The authors of Gated Graph ConvNet (GatedGCN) proposed using edge gates $\Mat{e}_{vu_i}^l \in \mathbb{R}^{C_{out} \times C_{in}}$ to control the flow of neighbors before aggregation \cite{bresson2017residual}. 

In the $l^{th}$ layer of GatedGCN module, the input node feature of the central node $v$ is denoted as $\Mat{h}_v^{l-1}$. The input node feature of the $i^{th}$ neighbor node $u_i$ is denoted as $\Mat{h}_{u_i}^{l-1}$. The input edge feature between $v$ and $u_i$ is denoted as $\hat{\Mat{e}}_{vu_i}^{l-1}$. The edge gates $\Mat{e}_{vu_i}^l$ are defined as:

\begin{equation}\label{con:e}
\Mat{e}_{vu_i}^l=\frac{\sigma(\hat{\Mat{e}}_{vu_i}^l)}{\sum_{j=1}^{N}{\sigma(\hat{\Mat{e}}_{vu_{j}}^l)}+\varepsilon},
\end{equation}
\begin{equation}
\hat{\Mat{e}}_{vu_i}^l=\relu\big(\bn(\Mat{A}^l\Mat{h}_{v}^{l-1}+\Mat{B}^l\Mat{h}_{u_i}^{l-1}+\Mat{C}^l\hat{\Mat{e}}_{vu_i}^{l-1})\big),
\label{con:GatedGCN_e}
\end{equation}
where $N=\left | \mathcal{N}(v) \right |$, and $\varepsilon$ is a small constant to keep numerical stability, $\Mat{A}^l,\Mat{B}^l,\Mat{C}^l \in \mathbb{R}^{C_{out} \times C_{in}}$, $\hat{e}_{vu_i}^0={e}_{vu_i}$. This edge gating mechanism not only fuses information from both nodes and edges but also passes it to the next layer, which makes it more powerful than GAT. By substituting Equation \eqref{con:GatedGCN_e} for $\hat{\Mat{e}}_{vu_i}^l$ in Equation~\eqref{con:e}
, the node update equation is defined as:

\begin{equation}
\Mat{h}_{v}^l=\relu\big(\bn({\Mat{U}^l\Mat{h}_{v}^{l-1}+\sum_{i=1}^{N}{\Mat{e}_{vu_i}^l \odot \Mat{V}^l\Mat{h}_{u_i}^{l-1}}})\big),
\end{equation}
where $\Mat{U}^l,\Mat{V}^l \in \mathbb{R}^{C_{out} \times C_{in}}$. 

\begin{figure*}[!ht]
  \centering
  \includegraphics[width=16cm,keepaspectratio]{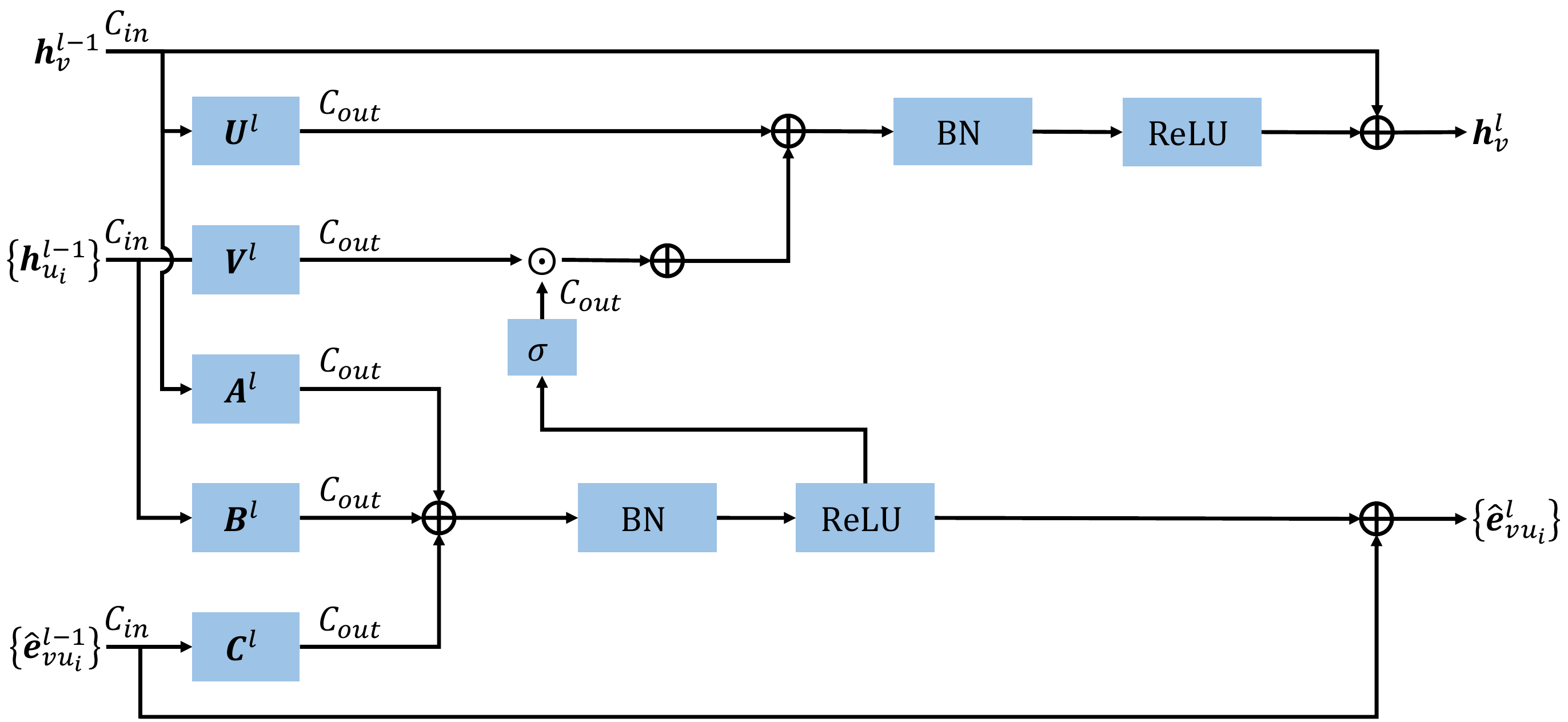}
  \caption{The architecture of GatedGCN module in the $l^{th}$ layer. $C_*$ indicates the dimension of feature vector. $\sigma$ is sigmoid activation function. $\bn$ is batch normalization.  \textcolor{Black}{$\protect\boldsymbol{\oplus}$} denotes the element-wise summation functions.} \label{fig:GatedGCN}
\end{figure*}

\begin{figure*}[!ht]
  \centering
  \includegraphics[width=\textwidth,keepaspectratio]{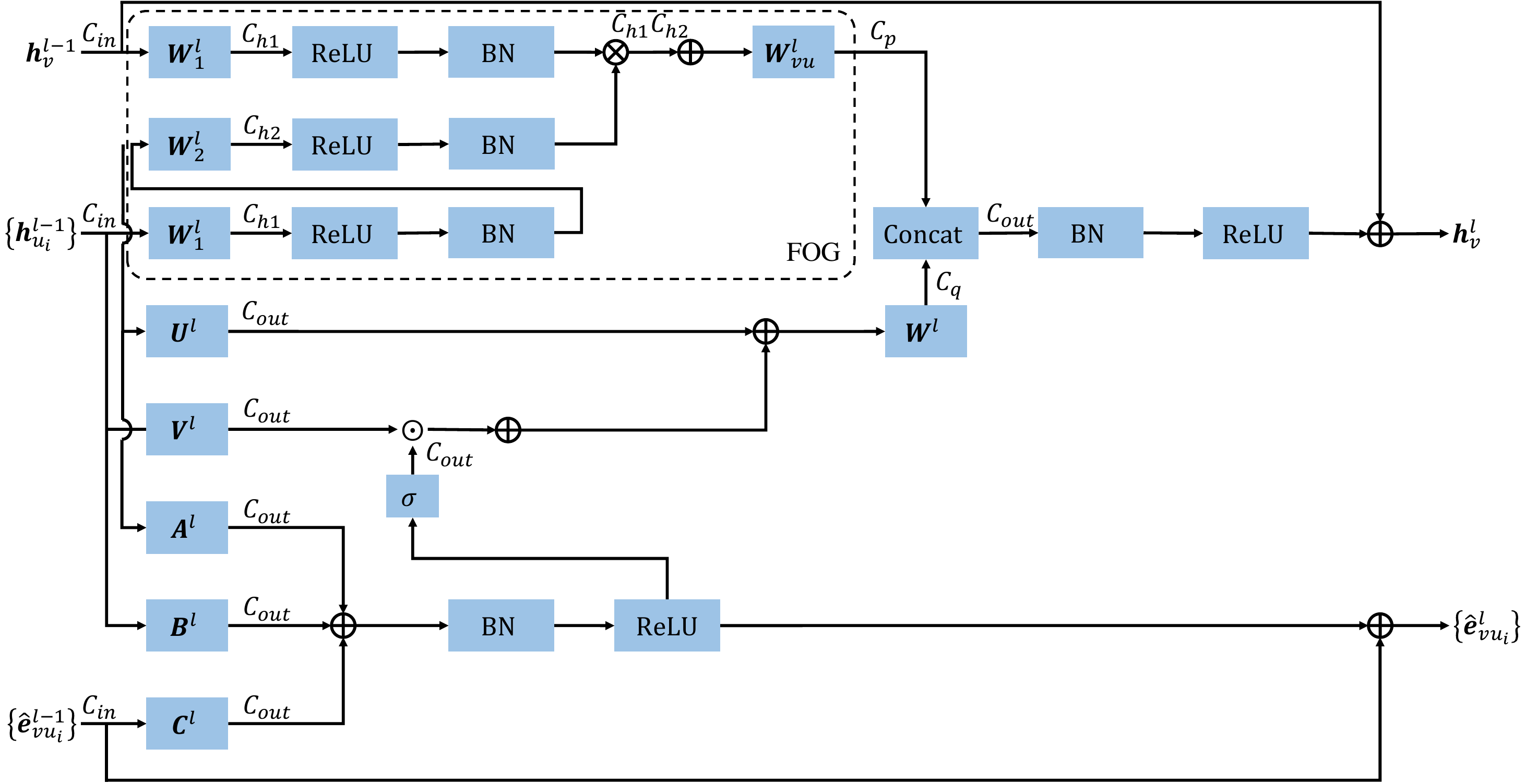}
  \caption{The architecture of GatedGCN+FOG module in the $l^{th}$ layer. $C_*$ indicates the dimension of feature vector. $\sigma$ is sigmoid activation function. $\concat$ is concatenation along the channel. $\bn$ is batch normalization.  \textcolor{Black}{$\protect\boldsymbol{\otimes}$} and \textcolor{Black}{$\protect\boldsymbol{\oplus}$} denote the Kronecker product and the element-wise summation functions, respectively.} \label{fig:GatedGCN_FOG}
\end{figure*}

After adding a linear mapping $\Mat{W}^l \in \mathbb{R}^{C_{q} \times C_{out}}$, the FOG module can be combined with GatedGCN as follows:

\begin{equation}
\Mat{h}_{v}^l=\relu\Big(\bn\big(\concat(\Mat{p}_{v}^l,\Mat{q}_{v}^l)\big)\Big),
\end{equation}

\begin{equation}
\begin{split}
\Mat{p}_{v}^l&=f_{FOG}(\Mat{h}_{v}^{l-1},\{\Mat{h}_{u_i}^{l-1}\}),    
\end{split}
\end{equation}

\begin{equation}
\begin{split}
\Mat{q}_{v}^l&=\Mat{W}^l({\Mat{U}^l\Mat{h}_{v}^{l-1}+\sum_{i=1}^{N}{\Mat{e}_{vu_i}^l \odot \Mat{V}^l\Mat{h}_{u_i}^{l-1}}}).        
\end{split}
\end{equation}
 The architecture of GatedGCN and GatedGCN+FOG module are illustrated as Figure \ref{fig:GatedGCN} and Figure \ref{fig:GatedGCN_FOG}, respectively.

\subsection{GIN with FOG} 
The Graph Isomorphism Network (GIN) proposed by \cite{xu2018powerful}  aims to achieve same representational power as Weisfeiler-Lehman (WL) graph isomorphism test \cite{leman1968reduction}.

In the $l^{th}$ layer of GIN, the input node feature of the central node $v$ is denoted as $\Mat{h}_v^{l-1}$. The input node feature of the $i^{th}$ neighbor node $u_i$ is denoted as $\Mat{h}_{u_i}^{l-1}$. The central node feature is updated as:

\begin{equation}\label{con:GIN}
\Mat{h}_{v}^l=\relu\Bigg(\bn\bigg({\Mat{U}^l\Big(\relu\big(\bn(\Mat{V}^l\hat{\Mat{h}}_{v}^{l}})\big)\Big)\bigg)\Bigg),
\end{equation}

\begin{equation}
\hat{\Mat{h}}_{v}^l=(1+\epsilon)\Mat{h}_v^{l-1}+\sum_{i=1}^{N}{\Mat{h}_{u_i}^{l-1}},
\end{equation}
where $\Mat{U}^l,\Mat{V}^l \in \mathbb{R}^{C_{out} \times C_{in}}$, $N=\left | \mathcal{N}(v) \right |$, and the learnable parameter $\epsilon$ is initialized as 0 at the beginning of training. After aggregation, feature vectors of central nodes are passed to a multilayer perceptron (MLP) which contains $N$ linear layers with the same dimension of outputs, $N-1$ batch normalization and $N-1$ $\relu$ activation functions. We set $N=2$ for all experiments where GIN modules are used.

\begin{figure*}[!ht]
  \centering
  \includegraphics[width=\textwidth,keepaspectratio]{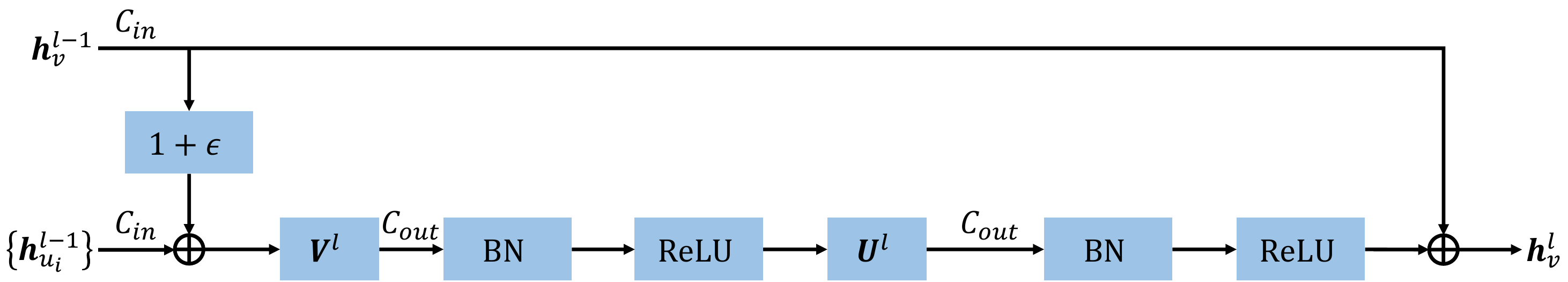}
  \caption{The architecture of GIN module in the $l^{th}$ layer. $C_*$ indicates the dimension of feature vector. $\epsilon$ is a learnable parameter. $\bn$ is batch normalization.  \textcolor{Black}{$\protect\boldsymbol{\otimes}$} and \textcolor{Black}{$\protect\boldsymbol{\oplus}$} denote the Kronecker product and the element-wise summation functions, respectively.} \label{fig:GIN}
\end{figure*}

\begin{figure*}[!ht]
  \centering
  \includegraphics[width=\textwidth,keepaspectratio]{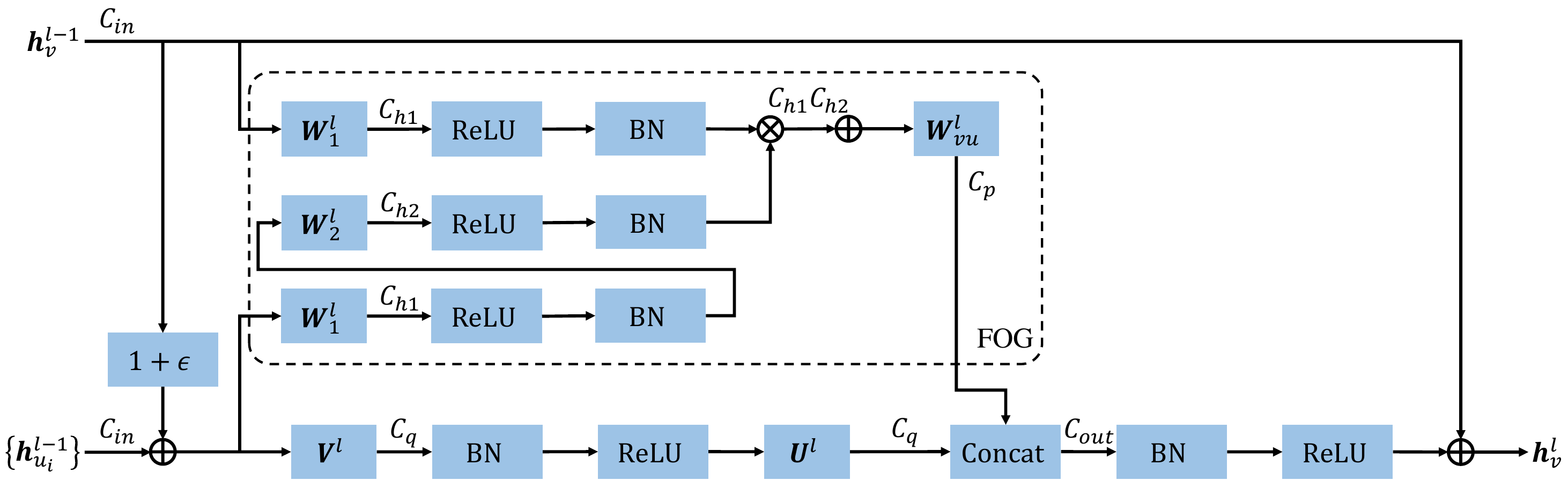}
  \caption{The architecture of GIN+FOG module in the $l^{th}$ layer. $C_*$ indicates the dimension of feature vector. $\epsilon$ is a learnable parameter. $\concat$ is concatenation along the channel. $\bn$ is batch normalization.  \textcolor{Black}{$\protect\boldsymbol{\otimes}$} and \textcolor{Black}{$\protect\boldsymbol{\oplus}$} denote the Kronecker product and the element-wise summation functions, respectively.} \label{fig:GIN_FOG}
\end{figure*}

By concatenating FOG to GIN, the Equation \ref{con:GIN} can be written as: 
\begin{equation}
{\Mat{h}}_{v}^l= \relu\Big(\bn\big(\concat(\Mat{p}_{v}^l,\Mat{q}_{v}^l)\big)\Big),
\end{equation}

\begin{equation}
\Mat{p}_{v}^l=f_{FOG}(\Mat{h}_{v}^{l-1},\{\Mat{h}_{u_i}^{l-1}\}), 
\end{equation}

\begin{equation}
\Mat{q}_{v}^l={\Mat{U}^l\Big(\relu\big(\bn(\Mat{V}^l\hat{\Mat{h}}_{v}^{l}})\big)\Big),
\end{equation}
where $\Mat{U}^l,\Mat{V}^l \in \mathbb{R}^{C_{q} \times C_{in}}$
The architecture of GIN and GIN+FOG module are illustrated as Figure \ref{fig:GIN} and Figure \ref{fig:GIN_FOG}, respectively.

\subsection{GraphSAGE with FOG}

Inspired by WL-test \cite{leman1968reduction}, GraphSAGE~\cite{hamilton2017inductive} incrementally aggregates information from further nodes through increasing the number of layers.

Given a central node $v$, a GraphSAGE module generates a new node feature $\Mat{h}_v^{l}$ for it by using its node feature $\Mat{h}_v^{l-1}$ and node features of neighbors, $\Mat{h}_u^{l-1}$, in the $l^{th}$ layer. This process can be formulated as~\cite{hamilton2017inductive}:

\begin{equation}
{\Mat{h}}_{v}^l=\relu\big(\Mat{U}^l\concat({\Mat{h}}_{v}^{l-1},\frac{1}{N}\sum_{i=1}^{N}{\Mat{h}_{u_i}^{l-1}})\big),
\label{con:GraphSAGE}
\end{equation}
where $\Mat{U}^l \in \mathbb{R}^{C_{out} \times C_{in}}$, and $N=\left | \mathcal{N}(v) \right |$. 

\begin{figure*}[!ht]
  \centering
  \includegraphics[width=12cm,keepaspectratio]{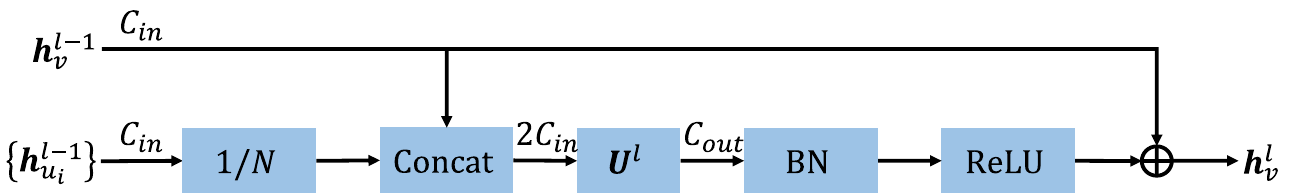}
  \caption{The architecture of GraphSAGE module in the $l^{th}$ layer. $C_*$ indicates the dimension of feature vector. $\bn$ is batch normalization.  \textcolor{Black}{$\protect\boldsymbol{\oplus}$} denotes the element-wise summation functions.} \label{fig:GraphSAGE}
\end{figure*}

\begin{figure*}[!ht]
  \centering
  \includegraphics[width=15cm,keepaspectratio]{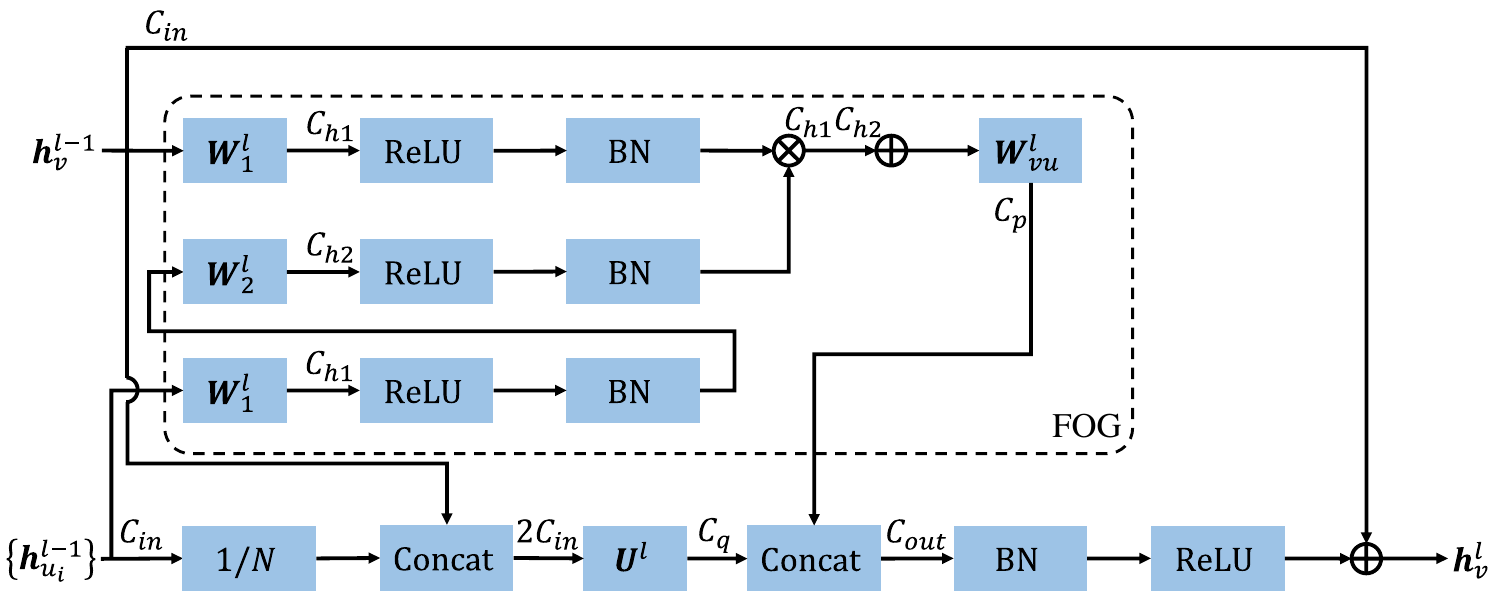}
  \caption{The architecture of GraphSAGE+FOG module in the $l^{th}$ layer. $C_*$ indicates the dimension of feature vector. $\concat$ is concatenation along the channel. $\bn$ is batch normalization.  \textcolor{Black}{$\protect\boldsymbol{\otimes}$} and \textcolor{Black}{$\protect\boldsymbol{\oplus}$} denote the Kronecker product and the element-wise summation functions, respectively.} \label{fig:GraphSAGE_FOG}
\end{figure*}

After introducing the FOG module into GraphSAGE, the $l^{th}$ layer is formulated as:

\begin{equation}
{\Mat{h}}_{v}^l= \relu\big(\concat(\Mat{p}_{v}^l,\Mat{q}_{v}^l)\big),
\label{con:Bi_GraphSAGE}
\end{equation}

\begin{equation}
\Mat{p}_{v}^l=f_{FOG}(\Mat{h}_{v}^{l-1},\{\Mat{h}_{u_i}^{l-1}\}), 
\end{equation}

\begin{equation}
\Mat{q}_{v}^l=\Mat{U}^l\concat({\Mat{h}}_{v}^{l-1},\frac{1}{N}\sum_{i=1}^{N}{\Mat{h}_{u_i}^{l-1}}),
\end{equation}
where $\Mat{U}^l \in \mathbb{R}^{{C_q} \times C_{in}}$. The architecture of GraphSAGE and GraphSAGE+FOG module are illustrated as Figure \ref{fig:GraphSAGE} and Figure \ref{fig:GraphSAGE_FOG}, respectively.

\subsection{GNN with FOG in the few-shot learning experiments}
Following to \cite{MessagePassing}, \cite{garcia2018fewshot} designs a GNN to aggregate information between query and its support set in the few-shot learning task.

In the $l^{th}$ layer of the GNN, the input node feature of the central node $v$ is denoted as $\Mat{h}_v^{l-1}$. The input node feature of the $i^{th}$ neighbor node $u_i$ is denoted as $\Mat{h}_{u_i}^{l-1}$. A coefficient $a_{vu_i}^l$ is learnt as:
\begin{equation}
a_{vu_i}^l=\softmax\big(\mlp(\left | {(\Mat{h}_v^{l-1}-\Mat{h}_{u_i}^{l-1})}\right | )\big).
\end{equation}
Then, the central node feature updating function is formulated as:
\begin{equation}\label{con:GNN}
{\Mat{h}}_{v}^l=\leakyrelu\Big(\bn\big(\Mat{V}^l\concat(\Mat{h_v}^{l-1},\sum_{i=1}^{N}{a_{vu_i}\Mat{h}_{u_i}^{l-1}})\big)\Big),
\end{equation}
where $\Mat{V}^l \in \mathbb{R}^{C_{out} \times 2C_{in}^l}$. 

\begin{figure*}[!ht]
  \centering
  \includegraphics[width=13cm,keepaspectratio]{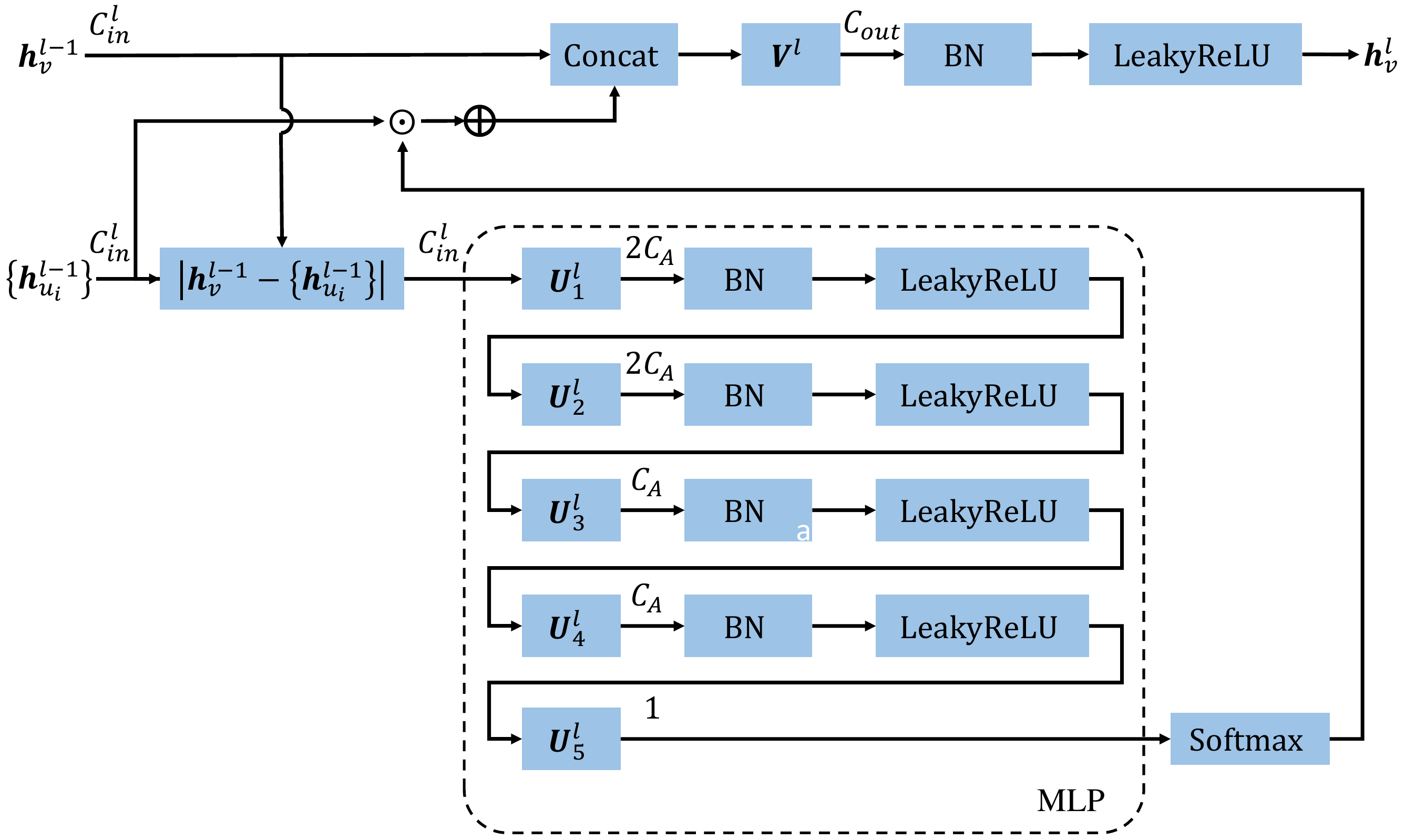}
  \caption{The architecture of GNN module in the $l^{th}$ layer. $C_*$ indicates the dimension of feature vector. $\concat$ is concatenation along the channel. $\bn$ is batch normalization.  \textcolor{Black}{$\protect\boldsymbol{\odot}$} denotes the Hadamard product. \textcolor{Black}{$\protect\boldsymbol{\oplus}$} denotes the element-wise summation functions.} \label{fig:Few_shot_GNN}
\end{figure*}

\begin{figure*}[!ht]
  \centering
  \includegraphics[width=13cm,keepaspectratio]{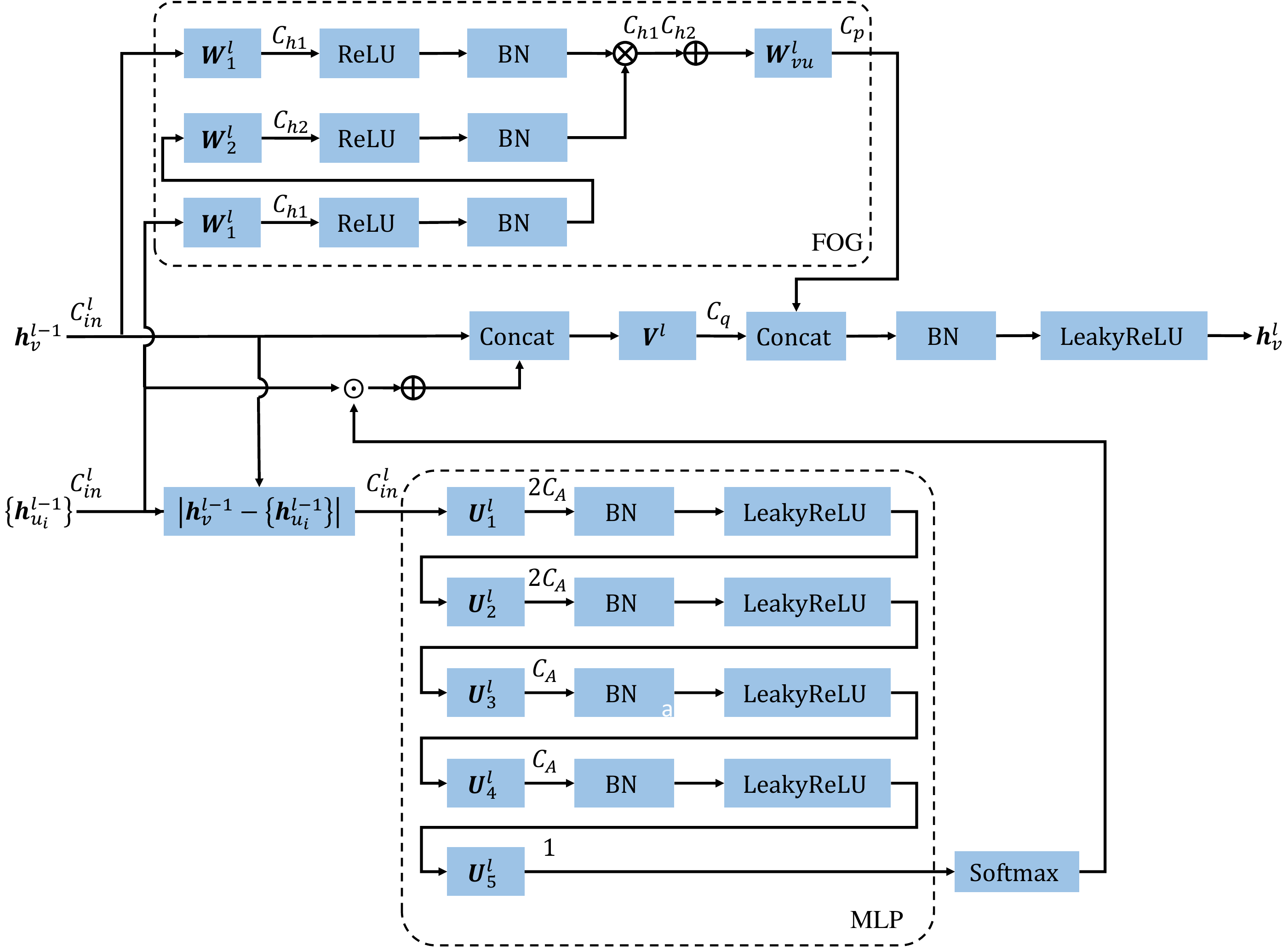}
  \caption{The architecture of GNN+FOG module in the $l^{th}$ layer. $C_*$ indicates the dimension of feature vector. $\concat$ is concatenation along the channel. $\bn$ is batch normalization.  \textcolor{Black}{$\protect\boldsymbol{\otimes}$} and \textcolor{Black}{$\protect\boldsymbol{\oplus}$} denote the Kronecker product and the element-wise summation functions, respectively. \textcolor{Black}{$\protect\boldsymbol{\odot}$} denotes the Hadamard product.} \label{fig:Few_shot_GNN_FOG}
\end{figure*}

After introducing the FOG module into the GNN, the Equation \ref{con:GNN} can be written as:
\begin{equation}
{\Mat{h}}_{v}^l= \leakyrelu\Big(\bn\big(\concat(\Mat{p}_{v}^l,\Mat{q}_{v}^l)\big)\Big),
\end{equation}

\begin{equation}
\Mat{p}_{v}^l=f_{FOG}(\Mat{h}_{v}^{l-1},\{\Mat{h}_{u_i}^{l-1}\}), 
\end{equation}

\begin{equation}
\Mat{q}_{v}^l=\Mat{V}^l\concat(\Mat{h_v}^{l-1},\sum_{i=1}^{N}{a_{vu_i}\Mat{h}_{u_i}^{l-1}}),
\end{equation}
where $\Mat{V}^l \in \mathbb{R}^{{C_q} \times C_{in}^l}$. The architecture of the GNN and GNN+FOG module are illustrated as Figure \ref{fig:Few_shot_GNN} and Figure \ref{fig:Few_shot_GNN_FOG}, respectively.

\section{Network Architecture Details}

\begin{figure*}[!ht]
  \centering
	\subfigure[]{\includegraphics[height=10cm, keepaspectratio]{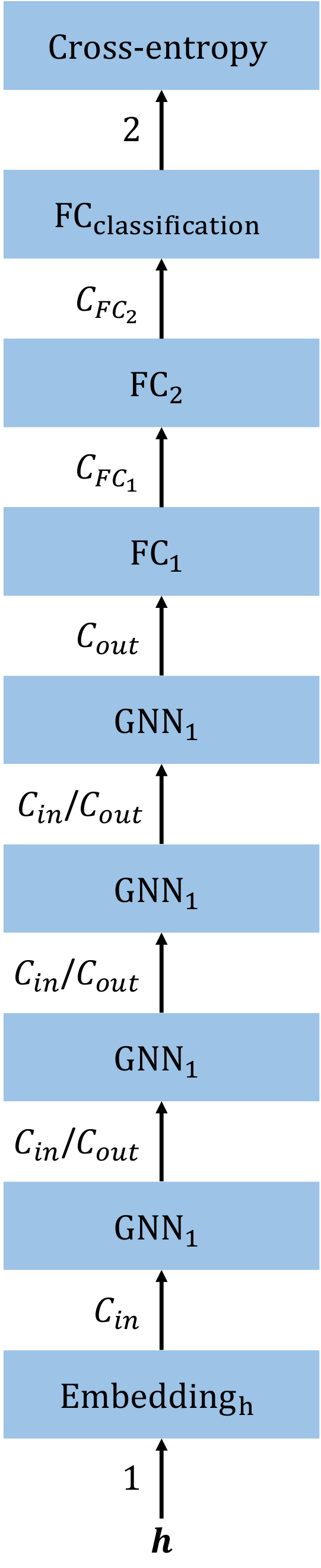}\label{PATTERN_GNN1}}
	\hfil
	\subfigure[]{\includegraphics[height=10cm, keepaspectratio]{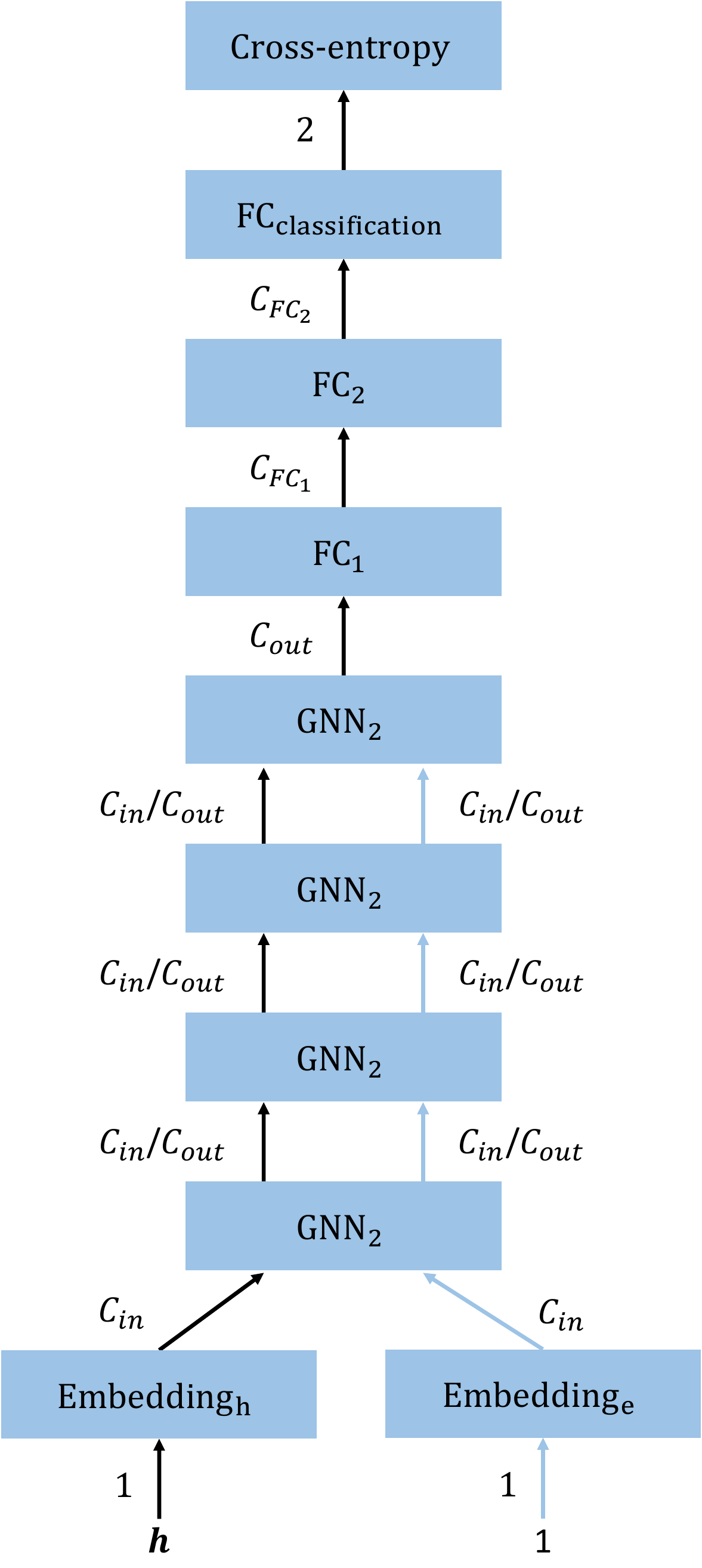}\label{PATTERN_GNN2}}
	\hfil
	\subfigure[]{\includegraphics[height=10cm, keepaspectratio]{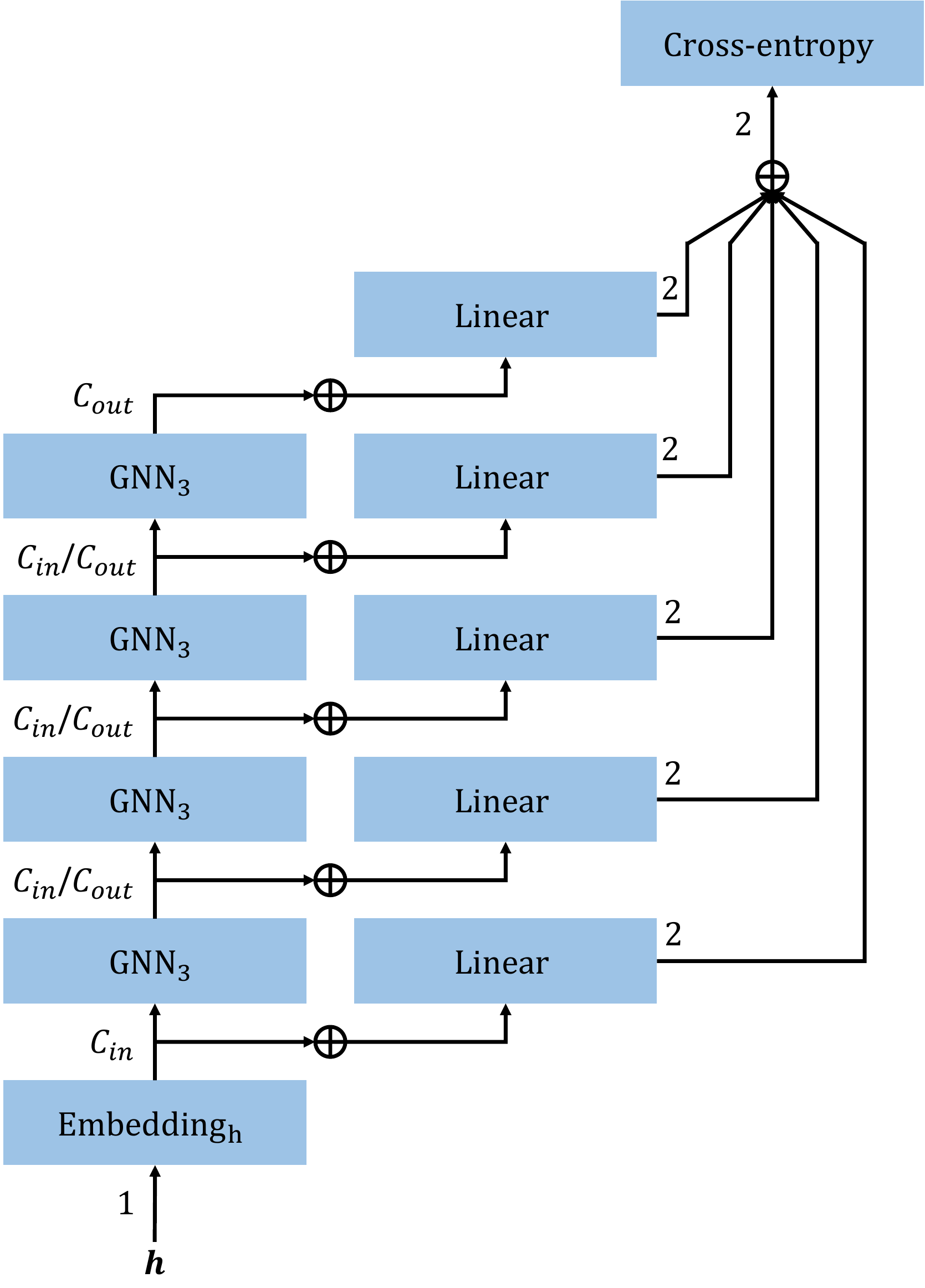}\label{PATTERN_GNN3}}
	\hfil

  \caption{The network architectures on SBM PATTERN. The \textbf{black} arrows indicate the flow of node features. The \textcolor{blue}{\textbf{blue}} arrows indicate the flow of edge features. The dimension of feature vectors are indicated beside the arrows. $\gnn_1$ can be switched among FOG, GCN, GCN+FOG, GAT, GAT+FOG, GraphSAGE, and GraphSAGE+FOG. $\gnn_2$ can be switched between GatedGCN and GatedGCN+FOG. $\gnn_3$ can be switched between GIN and GIN+FOG. \textcolor{Black}{$\protect\boldsymbol{\oplus}$} denotes the element-wise summation functions.} \label{fig:PATTERN}
\end{figure*}

\noindent{\textbf{Networks on SBM PATTERN.}} Each node in SBM PATTERN \cite{dwivedi2020benchmarking} is randomly assigned one label from $\{0,1,2\}$. The task of networks is identifying two connectivity patterns in each graph. Through $\embed_h$, the labels are embedded into node features. Especially, dummy edge features for GatedGCN and GatedGCN+FOG initialization are generated by $\embed_e$ through inputting $1$. The network architectures of GNNs and its FOG-equipped versions evaluated in this paper are illustrated as Figure \ref{fig:PATTERN}.

\begin{figure*}[!ht]
  \centering
	\subfigure[]{\includegraphics[height=10cm, keepaspectratio]{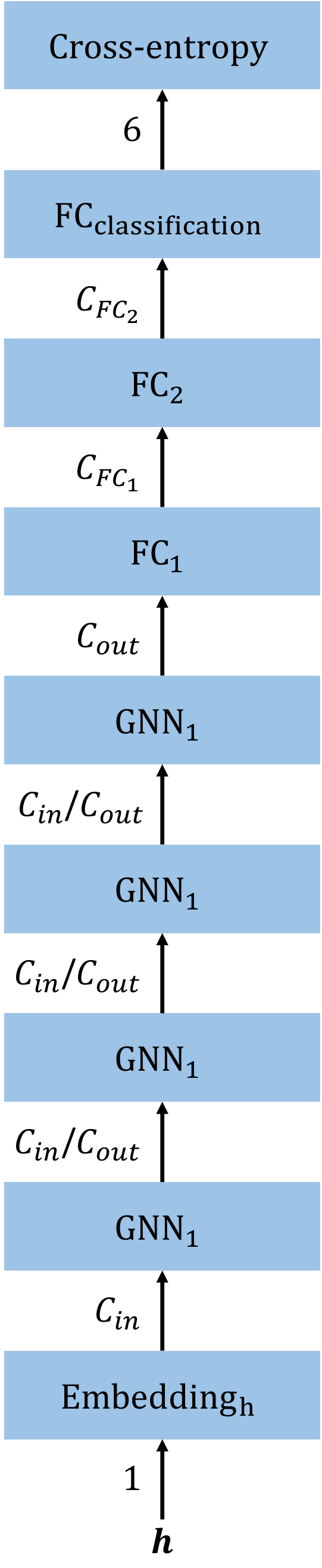}\label{CLUSTER_GNN1}}
	\hfil
	\subfigure[]{\includegraphics[height=10cm, keepaspectratio]{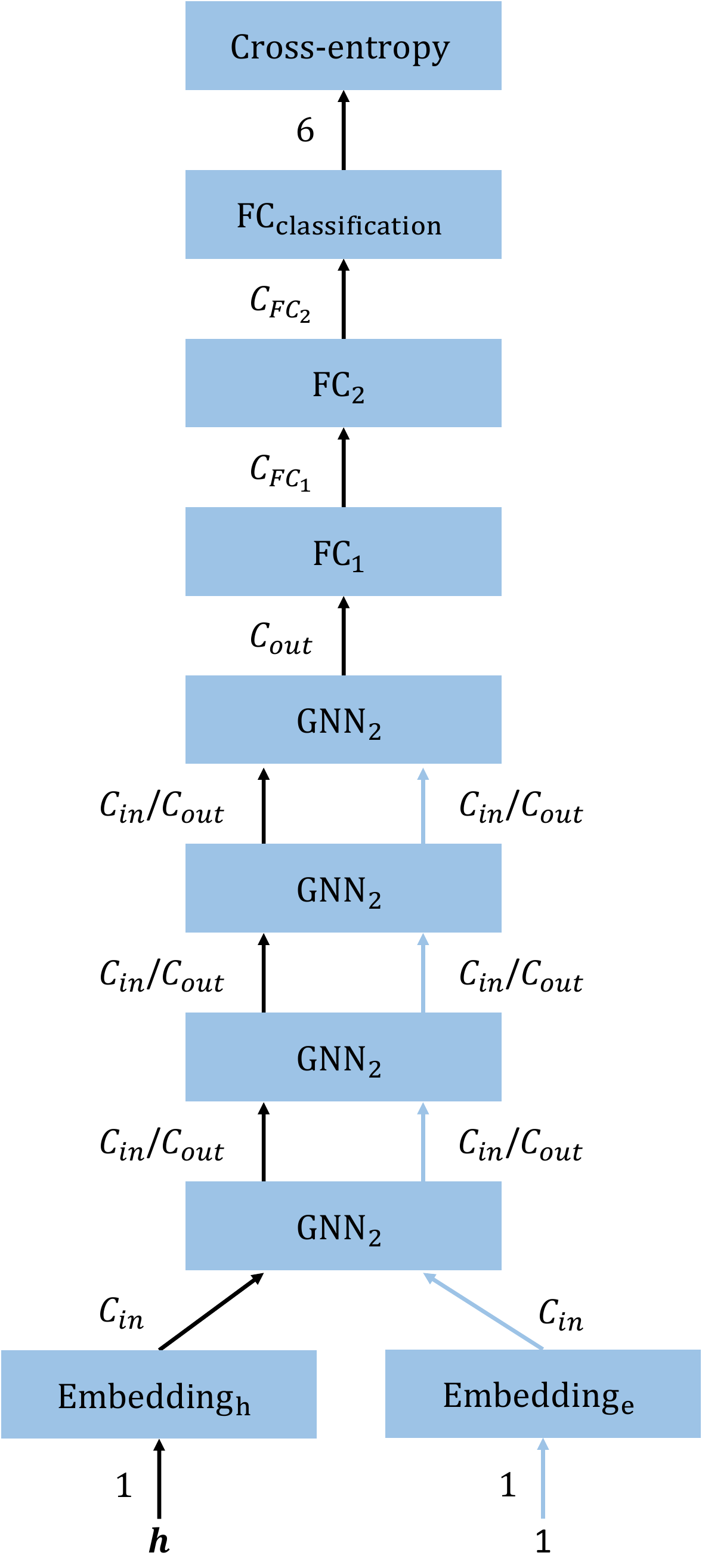}\label{CLUSTER_GNN2}}
	\hfil
	\subfigure[]{\includegraphics[height=10cm, keepaspectratio]{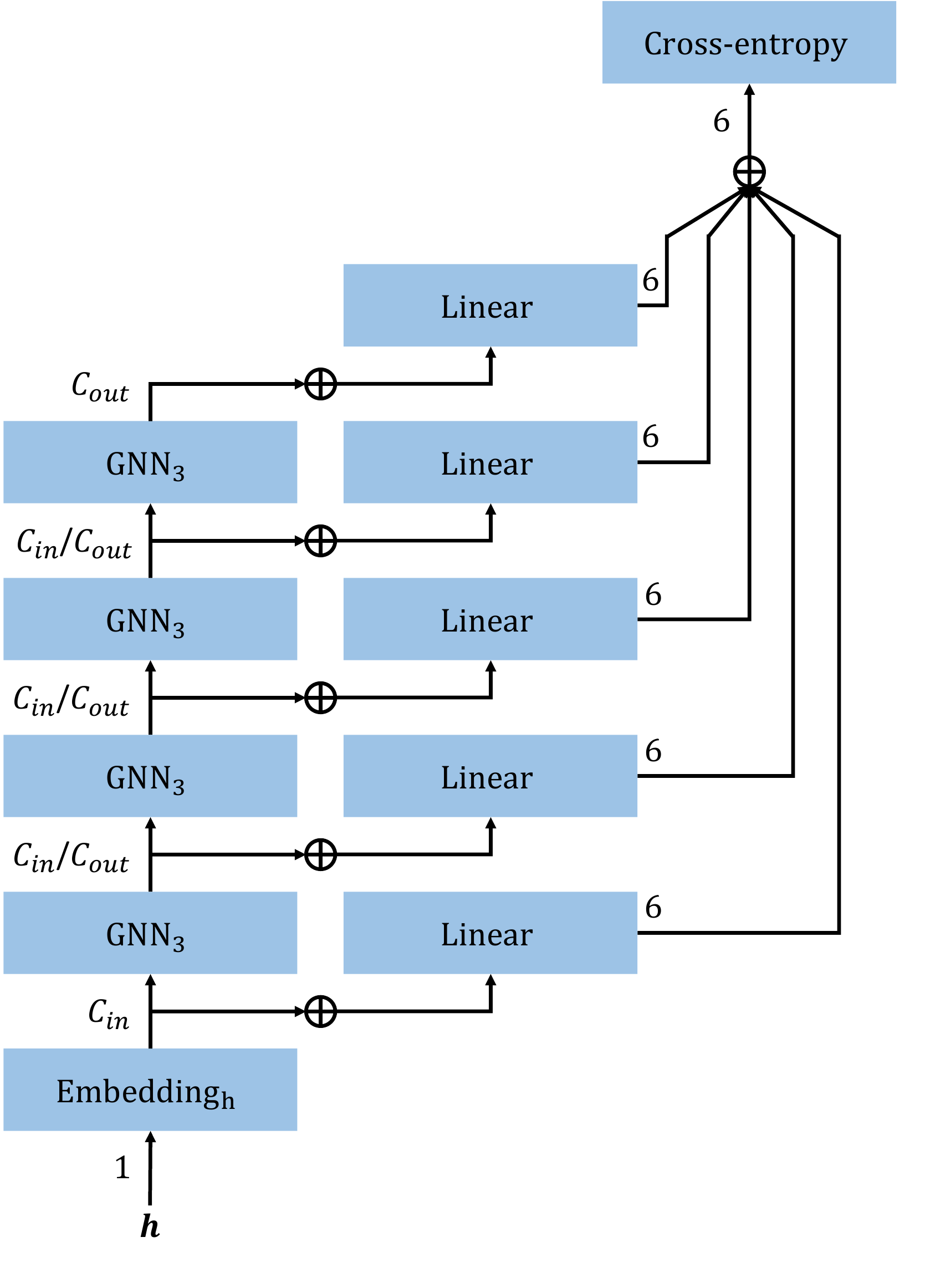}\label{CLUSTER_GNN3}}
	\hfil

  \caption{The network architectures on SBM CLUSTER. The \textbf{black} arrows indicate the flow of node features. The \textcolor{blue}{\textbf{blue}} arrows indicate the flow of edge features. The dimension of feature vectors are indicated beside the arrows. $\gnn_1$ can be switched among FOG, GCN, GCN+FOG, GAT, GAT+FOG, GraphSAGE, and GraphSAGE+FOG. $\gnn_2$ can be switched between GatedGCN and GatedGCN+FOG. $\gnn_3$ can be switched between GIN and GIN+FOG. \textcolor{Black}{$\protect\boldsymbol{\oplus}$} denotes the element-wise summation functions.} \label{fig:CLUSTER}
\end{figure*}

\noindent{\textbf{Networks on SBM CLUSTER.}} There are 6 communities in each graph in SBM CLUSTER \cite{dwivedi2020benchmarking}. Only one node in each community is randomly picked and assigned with one label from $\{1,2,3,4,5,6\}$. The remaining nodes are assigned with $0$. The task of networks is clustering these 6 types of community in each graph. Through $\embed_h$, the labels are embedded into node features. Especially, dummy edge features for GatedGCN and GatedGCN+FOG initialization are generated by $\embed_e$ through inputting $1$. The network architectures of GNNs and its FOG-equipped versions evaluated in this paper are illustrated as Figure \ref{fig:CLUSTER}.

\begin{figure*}[!ht]
  \centering
	\subfigure[]{\includegraphics[height=8cm, keepaspectratio]{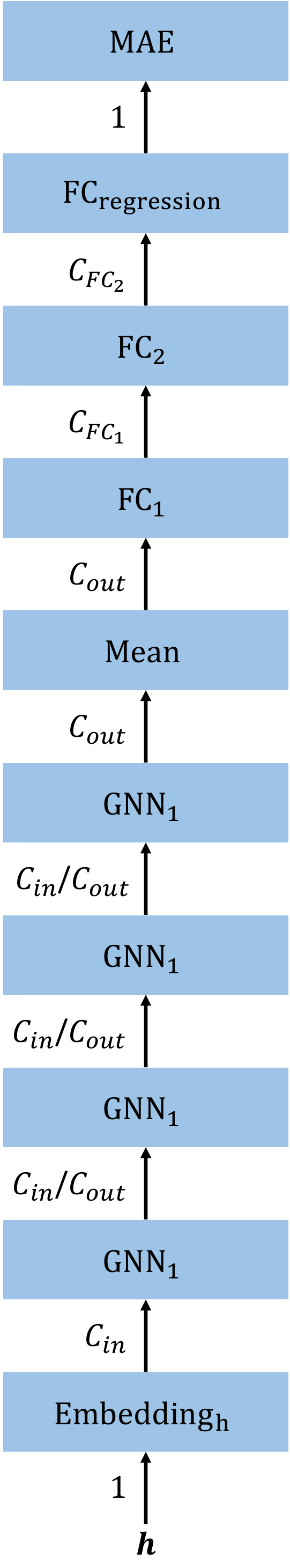}\label{ZINC_GNN1}}
	\hfil
	\subfigure[]{\includegraphics[height=8cm, keepaspectratio]{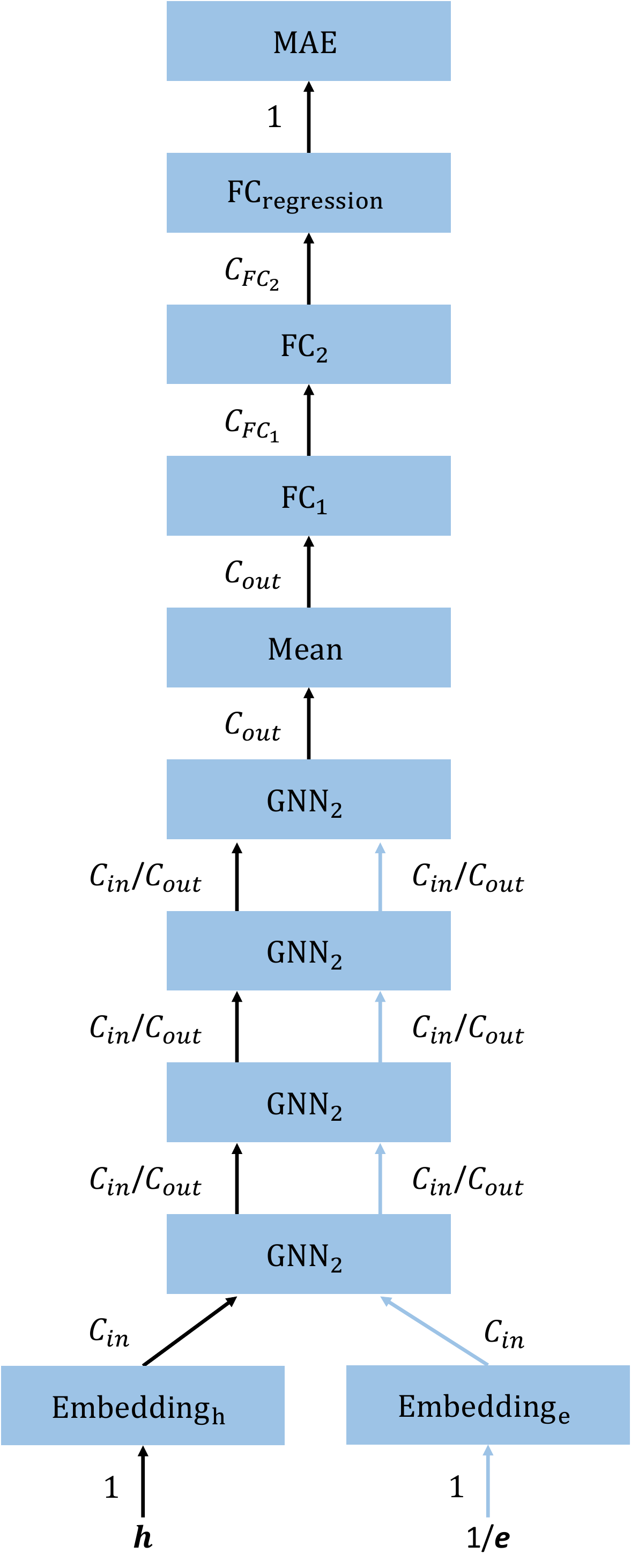}\label{ZINC_GNN2}}
	\hfil
	\subfigure[]{\includegraphics[height=8cm, keepaspectratio]{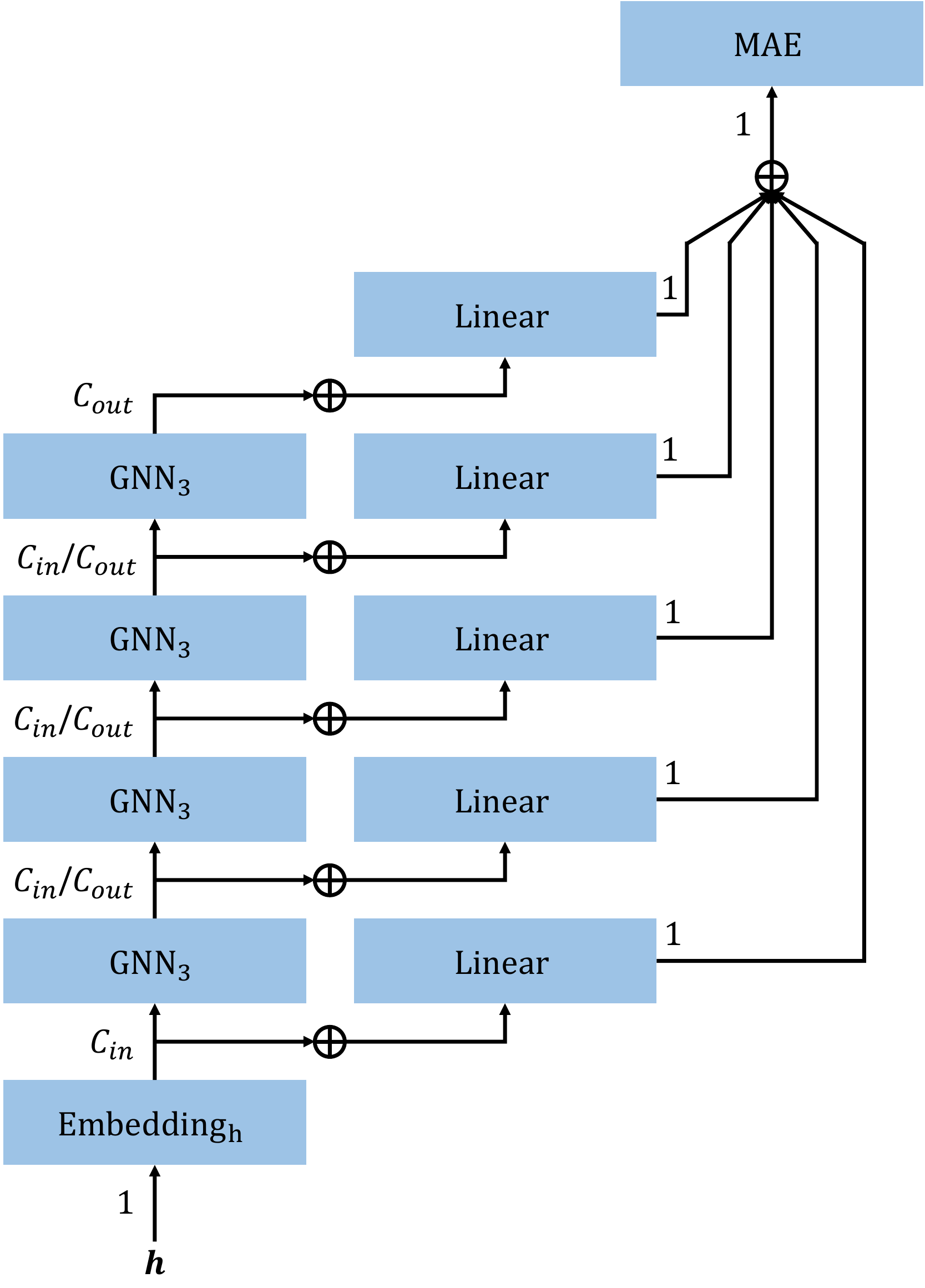}\label{ZINC_GNN3}}
	\hfil

  \caption{The network architectures on ZINC. The \textbf{black} arrows indicate the flow of node features. The \textcolor{blue}{\textbf{blue}} arrows indicate the flow of edge features. The dimension of feature vectors are indicated beside the arrows. $\gnn_1$ can be switched among FOG, GCN, GCN+FOG, GAT, GAT+FOG, GraphSAGE, and GraphSAGE+FOG. $\gnn_2$ can be switched between GatedGCN and GatedGCN+FOG. $\gnn_3$ can be switched between GIN and GIN+FOG. \textcolor{Black}{$\protect\boldsymbol{\oplus}$} denotes the element-wise summation functions.} \label{fig:ZINC}
\end{figure*}

\noindent{\textbf{Networks on ZINC.}} ZINC \cite{ZINC_data} contains 28 types of atoms and 4 types of bonds. The task of networks is regressing the constrained solubility of each graph (molecular). Through $\embed_h$ and $\embed_{e}$, the atoms and and bonds are embedded into node features and edge features, respectively. Especially, types of bonds are ignored and replaced with $1$ for GatedGCN, GatedGCN+FOG, and GatedGCN+GIN. For experiments on ZINC without node features, the types of atoms are ingnored and replaced with $1$. The network architectures of GNNs and its FOG-equipped versions evaluated in this paper are illustrated as Figure \ref{fig:ZINC}.

\begin{figure*}[!ht]
  \centering
	\subfigure[]{\includegraphics[height=10cm, keepaspectratio]{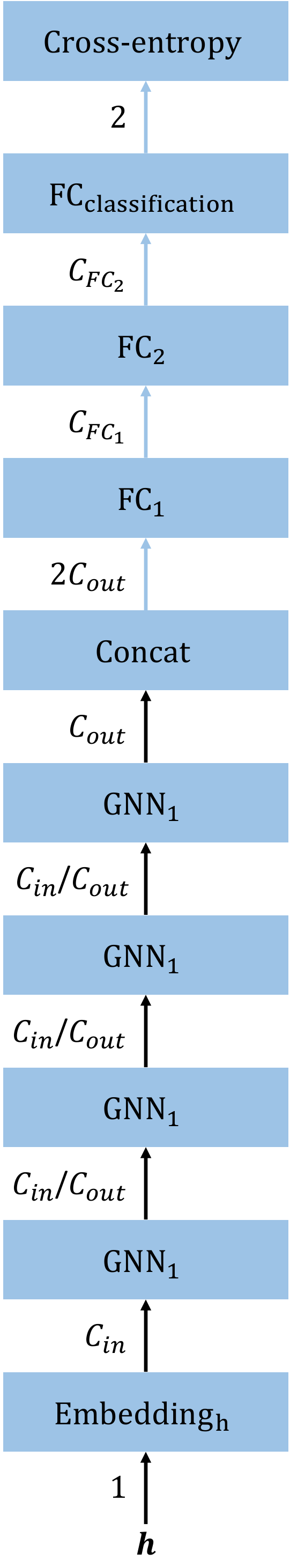}\label{TSP_GNN1}}
	\hfil
	\subfigure[]{\includegraphics[height=10cm, keepaspectratio]{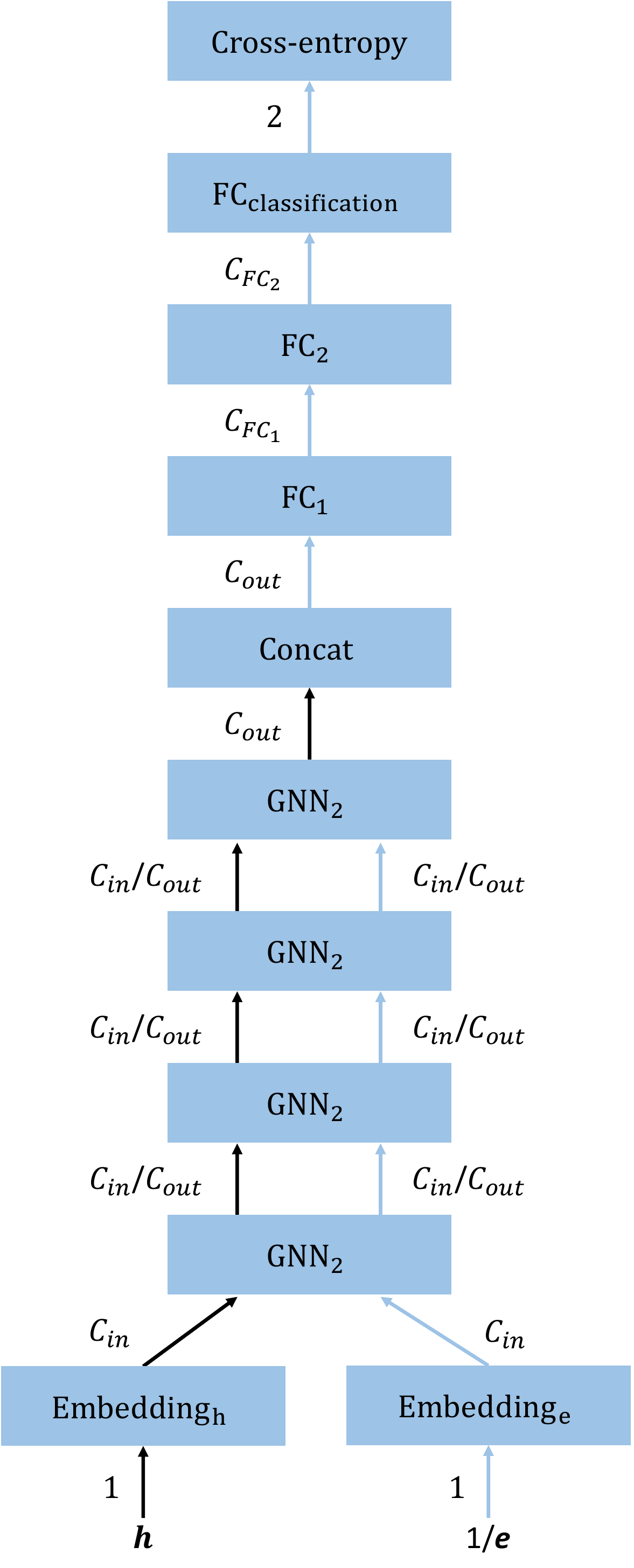}\label{TSP_GNN2}}
	\hfil
	\subfigure[]{\includegraphics[height=10cm, keepaspectratio]{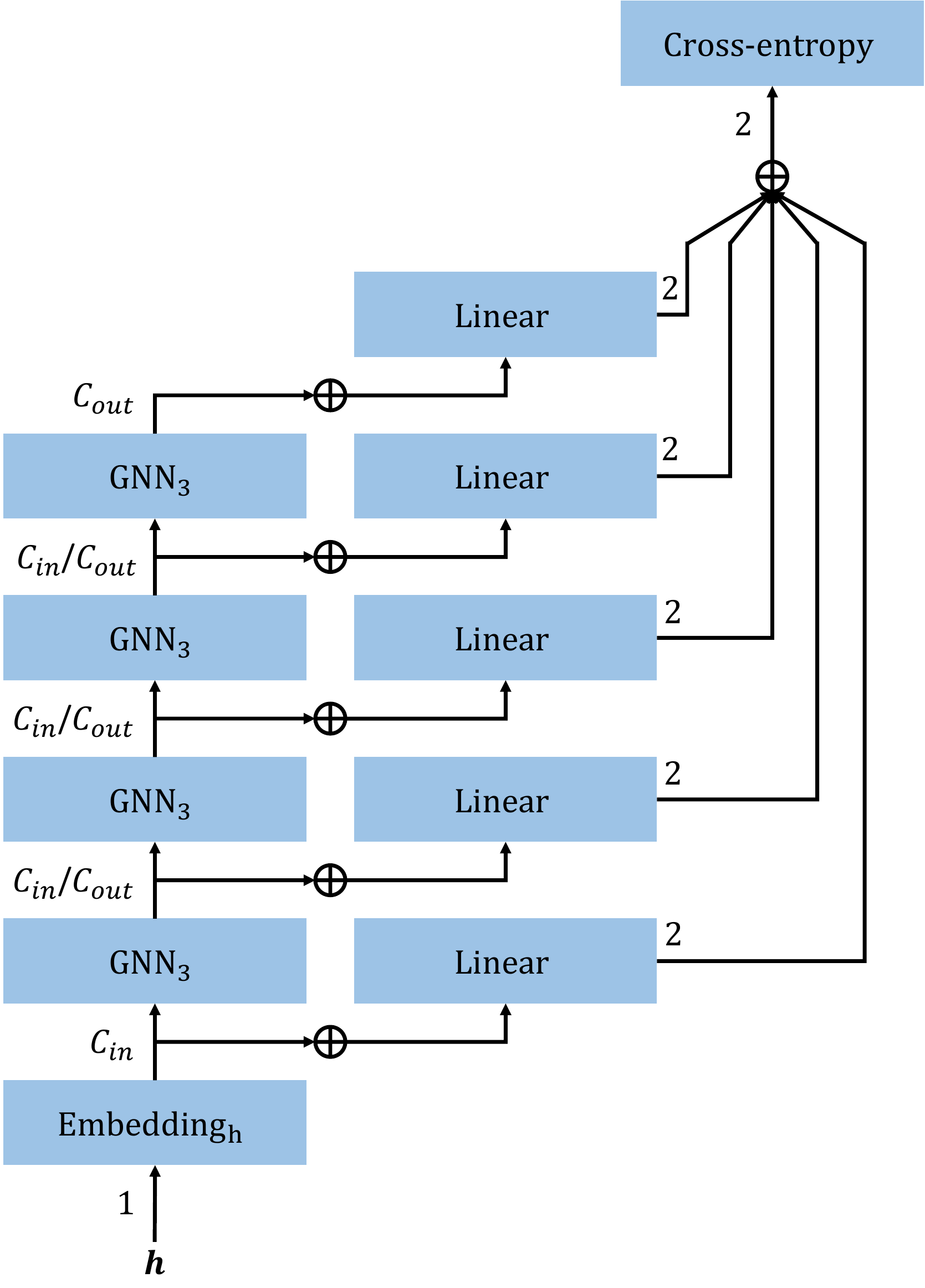}\label{TSP_GNN3}}
	\hfil

  \caption{The network architectures on TSP. The \textbf{black} arrows indicate the flow of node features. The \textcolor{blue}{\textbf{blue}} arrows indicate the flow of edge features. The dimension of feature vectors are indicated beside the arrows. $\gnn_1$ can be switched among FOG, GCN, GCN+FOG, GAT, GAT+FOG, GraphSAGE, and GraphSAGE+FOG. $\gnn_2$ can be switched between GatedGCN and GatedGCN+FOG. $\gnn_3$ can be switched between GIN and GIN+FOG. \textcolor{Black}{$\protect\boldsymbol{\oplus}$} denotes the element-wise summation functions.} \label{fig:TSP}
\end{figure*}

\noindent{\textbf{Networks on TSP.}} Each node in TSP \cite{dwivedi2020benchmarking} contains a two-dimensional coordinate as a node feature. The edge features are euclidean distances between nodes. The task of networks is predicting whether each edge is belong to the TSP tour given by Concorde TSP Solver of not. Through $\embed_h$ and $\embed_{e}$, the coordinates and and distances are embedded into node features and edge features for inputting GNN modules, respectively. Especially, the distances are ignored and replaced with $1$ for GatedGCN and GatedGCN+FOG. After the last GNN module, edge features are generated by concatenated node features belonging to it. The network architectures of GNNs and its FOG-equipped versions evaluated in this paper are illustrated as Figure \ref{fig:TSP}.

\begin{figure*}[!ht]
  \centering
	\subfigure[]{\includegraphics[height=8cm, keepaspectratio]{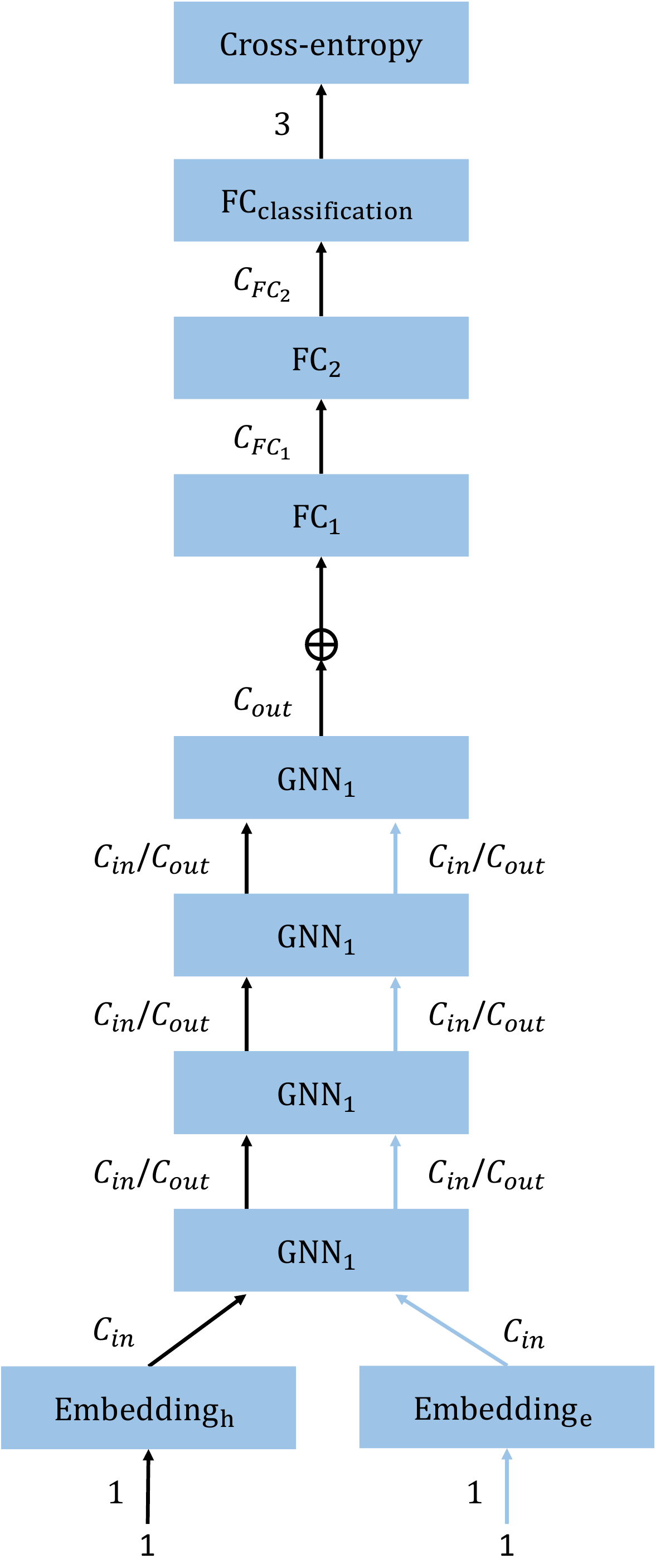}\label{IMDB_M_GNN1}}
	\hfil
	\subfigure[]{\includegraphics[height=8cm, keepaspectratio]{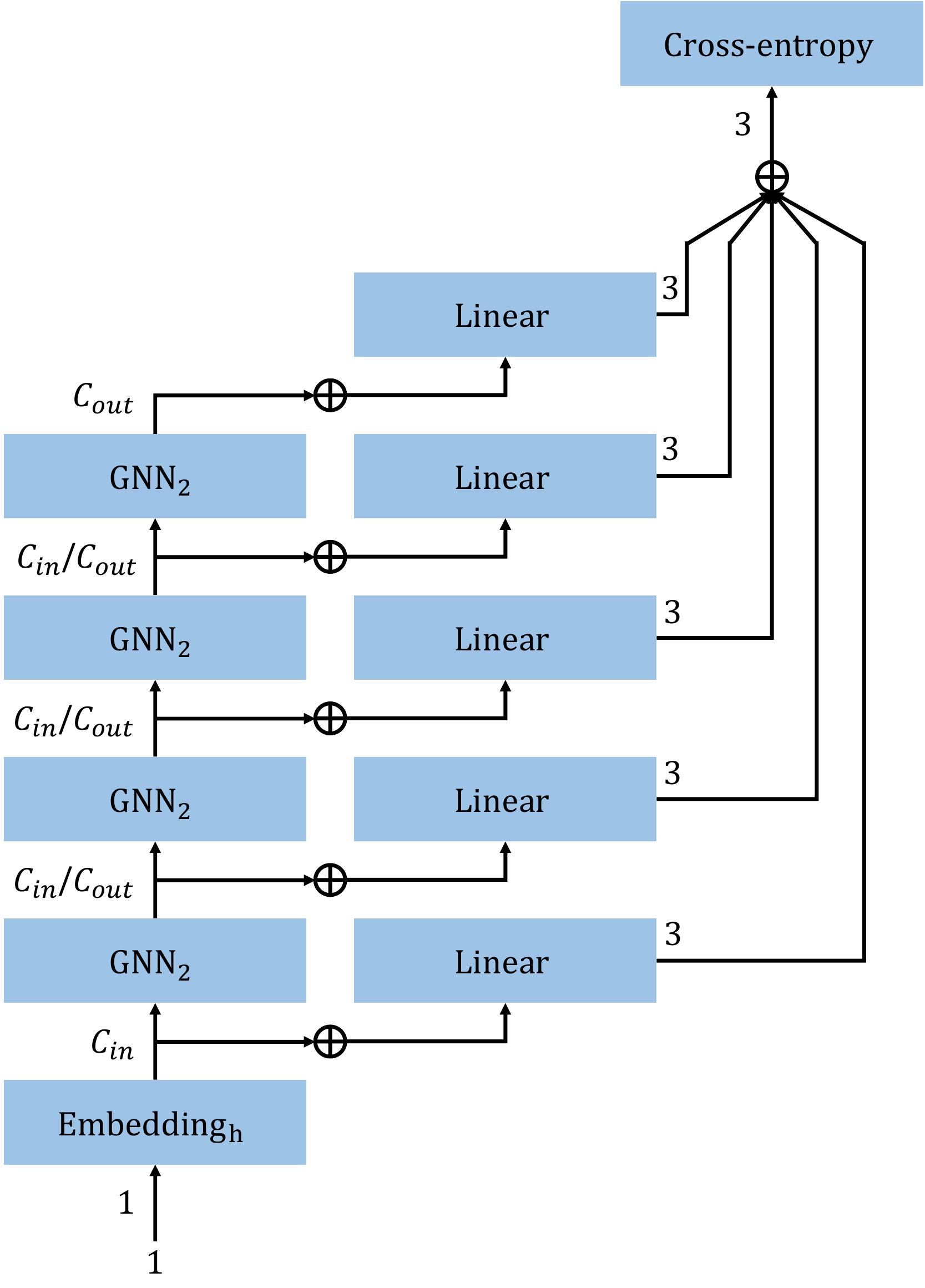}\label{IMDB_M_GNN2}}
	\hfil
	\subfigure[]{\includegraphics[height=8cm, keepaspectratio]{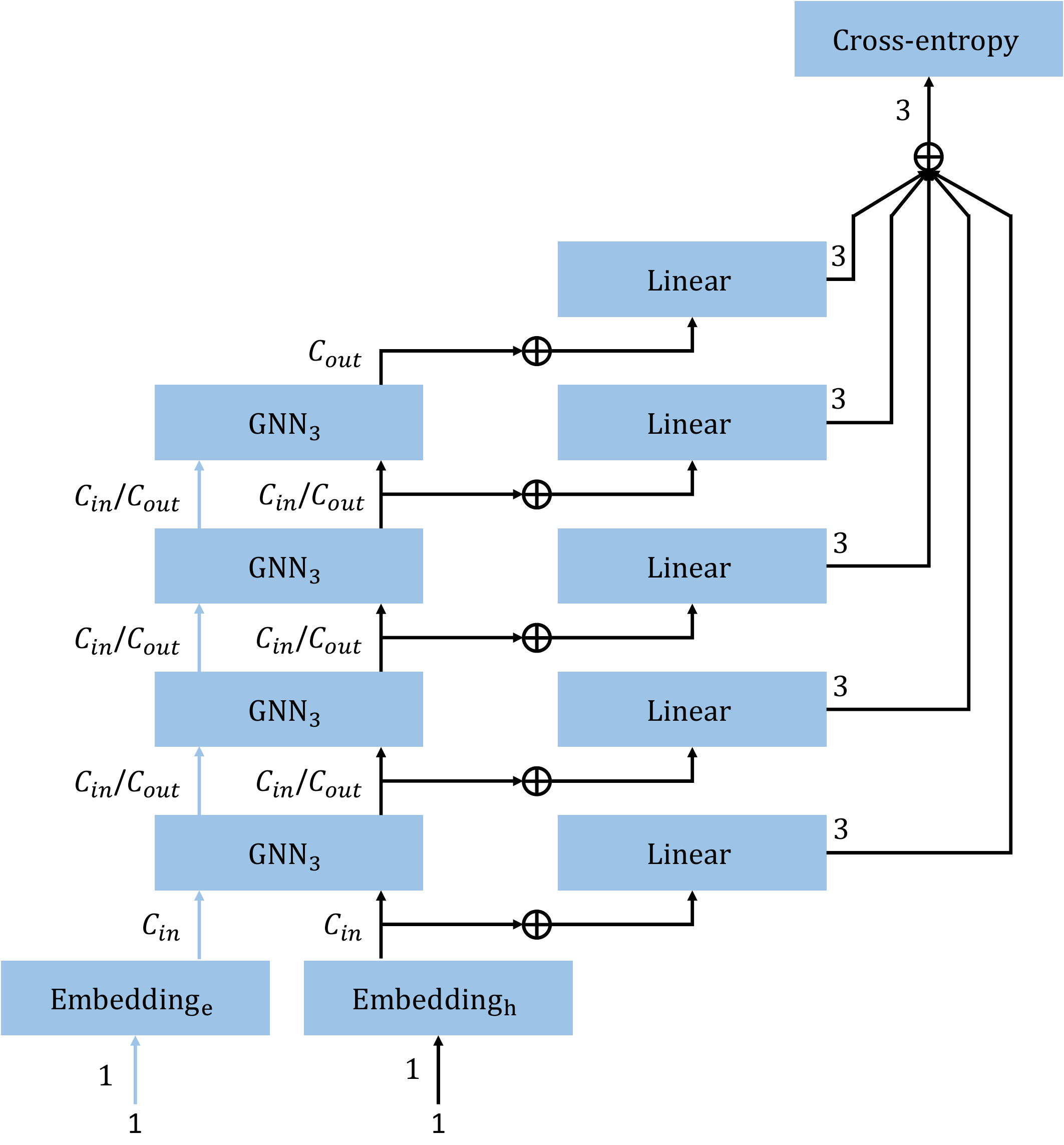}\label{IMDB_M_GNN3}}
	\hfil

  \caption{The network architectures on IMDB-MULTI. The \textbf{black} arrows indicate the flow of node features. The \textcolor{blue}{\textbf{blue}} arrows indicate the flow of edge features. The dimension of feature vectors are indicated beside the arrows. $\gnn_1$ can be switched between GatedGCN and GatedGCN+FOG. $\gnn_2$ can be switched between GIN and GIN+FOG. $\gnn_3$ represents GatedGCN+GIN. \textcolor{Black}{$\protect\boldsymbol{\oplus}$} denotes the element-wise summation functions.} \label{fig:IMDB_M}
\end{figure*}

\noindent{\textbf{Networks on IMDB-MULTI.}} Nodes in IMDB-MULTI \cite{DeepGraphKernels} represent actors/actresses and the values are identically set as $1$. The task of networks is classifying the genre of each graph representing from Comedy, Romance
and Sci-Fi. For the purpose of fair comparison, an embedding layer, $\embed_h$, is used to embed the original inputs into node feature vectors for all networks. Especially, dummy edge features for GatedGCN GatedGCN+FOG, and GatedGCN+GIN initialization are generated by $\embed_e$ through inputting $1$.  The network architectures of GNNs and its FOG-equipped versions evaluated in this paper are illustrated as Figure \ref{fig:IMDB_M}.

\begin{figure*}[!ht]
  \centering
	\subfigure[]{\includegraphics[height=12cm, keepaspectratio]{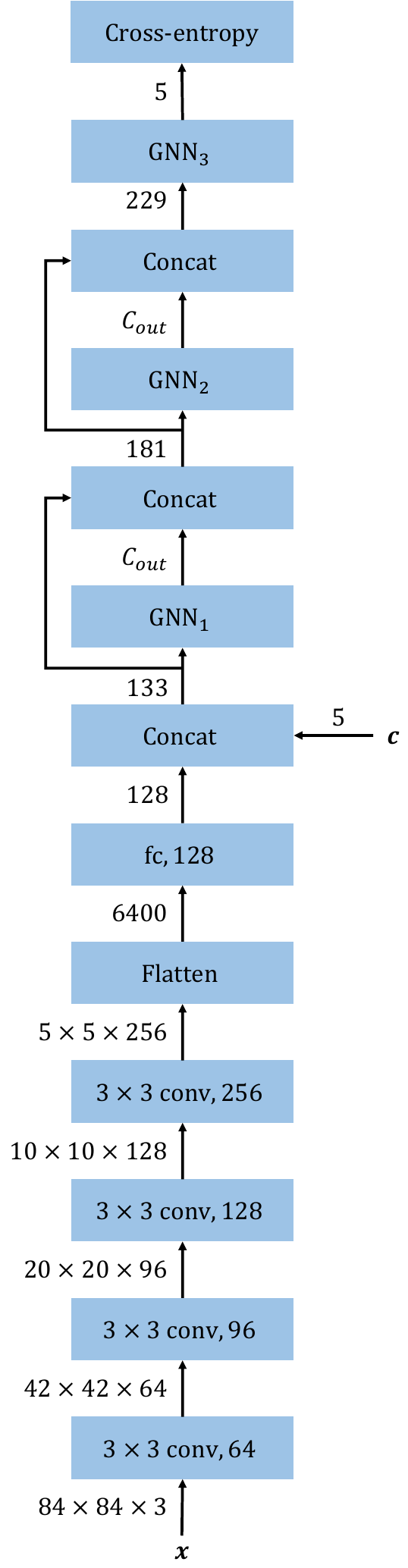}\label{Few_shot}}

  \caption{The network architectures on miniImageNet. The dimension of features are indicated beside the arrows. $\gnn_1$ and $\gnn_2$ can be switched among FOG, GNN, GNN+FOG. $\gnn_3$ represents the GNN originally used in this task.} \label{fig:miniImageNet}
\end{figure*}

\noindent{\textbf{Networks on miniImageNet.}} Node features in the graph formed by each task of few-shot learning are extracted by a 5-lay CNN. The task is classifying the query node among the support set. Especially, each node feature vector generated by the CNN is concatenated to a one-hot code $\Mat{c}$ which indicates its label in the graph. The code of query is $\Mat{0}$. The network architecture of GNN and its FOG-equipped versions evaluated in this paper are illustracted as Figure \ref{fig:miniImageNet}.

\subsection{Hyperparameter Settings}
Hyperparameter settings for all datasets and networks evaluated in this paper are listed in Table \ref{table:hyperparam_PATTERN}, \ref{table:hyperparam_CLUSTER}, \ref{table:hyperparam_ZINC}, \ref{table:hyperparam_ZINC_wo_nodeFeat},  \ref{table:hyperparam_TSP}, \ref{table:hyperparam_IMDBMULTI}, \ref{table:hyperparam_5w1s}, and \ref{table:hyperparam_5w5s}

\begin{table*}[!ht]
\caption{Hyperparameter Settings of all experiments on on SBM PATTERN. $\#Param$ is the number of training parameters; $L$ is the number of GNN module layers; $C_*$ are channel dimensions in Figure \ref{fig:FOG}, \ref{fig:GCN}, \ref{fig:GCN_FOG}, \ref{fig:GAT}, \ref{fig:GAT_FOG}, \ref{fig:GatedGCN}, \ref{fig:GatedGCN_FOG}, \ref{fig:GIN}, \ref{fig:GIN_FOG} , \ref{fig:GraphSAGE}, and \ref{fig:GraphSAGE_FOG}; $Init$ $lr$ is the initial learning rate; $wd$ is the weight decay; $Patience$ is the decay patience; $Min$ $lr$ is the stopping learning rate. Learning rate reduce factor and batch size are set to $0.5$ and $128$ for all experiments, respectively.}
\begin{center}
\scalebox{0.78}{
\begin{threeparttable}
\begin{tabular}{l|c|cccccccl|cccc}
\toprule[2pt]
\multirow{2}{*}{Model} & \multirow{2}{*}{\#Param} & \multicolumn{8}{c|}{Hyperparameters} & \multicolumn{3}{c}{Learning}\\
\cline{3-14}
&&$L$&$C_{in}/C_{out}$&$C_{h1}$&$C_{h2}$&$C_p/C_q$&$C_{FC_1}$&$C_{FC_2}$&Other&Init lr& \multicolumn{1}{c|}{wd}&Patience & Min lr \\
\hline
FOG & 99,046 & 4 & 144 & 16 & 8 & - & 72 & 36 & - & 5e-3 & \multicolumn{1}{c|}{1e-3}& \multirow{13}{*}{\shortstack[c]{5}} & \multirow{13}{*}{\shortstack[c]{1e-5}}\\
\cline{1-12}
GCN & 100,923 & 4 & 146 & - && - & 73 & 36 & - & 1e-3 & \multicolumn{1}{c|}{0} & &\\
\cline{1-12}
GCN+FOG & 101,026 & 4 & 160 & 12 & 6 & 80 & 80 & 40 & - &  5e-3  & \multicolumn{1}{c|}{1e-3} & &\\
\cline{1-12}

\multirow{2}{*}{\shortstack[l]{GAT}} & \multirow{2}{*}{\shortstack[c]{109,936}} & \multirow{2}{*}{\shortstack[c]{4}} & \multirow{2}{*}{\shortstack[c]{152}} & \multirow{2}{*}{\shortstack[c]{-}} & \multirow{2}{*}{\shortstack[c]{-}} & \multirow{2}{*}{\shortstack[c]{-}} & \multirow{2}{*}{\shortstack[c]{76}} & \multirow{2}{*}{\shortstack[c]{38}} & $K=8$ for the $1^{st}$ to $3^{rd}$ module; & \multirow{2}{*}{\shortstack[c]{1e-3}}  & \multicolumn{1}{c|}{\multirow{2}{*}{\shortstack[c]{0}}}  &  & \\
&  &  &  &  &  & & &  & $K=1$ for the $4^{th}$ module; &  & \multicolumn{1}{c|}{} & & \\
\cline{1-12}
\multirow{2}{*}{\shortstack[l]{GAT+FOG}} & \multirow{2}{*}{\shortstack[c]{101,346}} & \multirow{2}{*}{\shortstack[c]{4}} & \multirow{2}{*}{\shortstack[c]{160}} & \multirow{2}{*}{\shortstack[c]{12}} &\multirow{2}{*}{\shortstack[c]{6}} & \multirow{2}{*}{\shortstack[c]{80}} & \multirow{2}{*}{\shortstack[c]{80}} & \multirow{2}{*}{\shortstack[c]{40}} & $K=8$ for the $1^{st}$ to $3^{rd}$ module; & \multirow{2}{*}{\shortstack[c]{5e-3}}  &  \multicolumn{1}{c|}{\multirow{2}{*}{\shortstack[c]{1e-3}}} & &\\
&  &  &  &  &  & & &  & $K=1$ for the $4^{th}$ module; &  & \multicolumn{1}{c|}{} & &\\
\cline{1-12}

GatedGCN & 104,003 & 4 & 70 & - & - & - & 35 & 17 & - & 1e-3  & \multicolumn{1}{c|}{0}  & & \\
\cline{1-12}
GatedGCN+FOG & 102,050 & 4 & 64  & 8 & 4 & 32 & 32 & 16 & - & 1e-2   & \multicolumn{1}{c|}{1e-3} & &\\
\cline{1-12}
GIN & 100,884 & 4 & 110 & - & - & - & - & - & - & 1e-3  & \multicolumn{1}{c|}{0}  & & \\
\cline{1-12}
GIN+FOG & 99,234 & 4 & 148  & 12 & 6 & 74 & - & - & - & 5e-3 & \multicolumn{1}{c|}{1e-3}& & \\
\cline{1-12}
GraphSAGE & 101,739 & 4 & 89 & - & - & - & 44 & 22 & - & 1e-3  & \multicolumn{1}{c|}{0}  & & \\
\cline{1-12}
GraphSAGE+FOG & 95,679 & 4 & 98  & 9 & 4 & 49 & 49 & 24 & - & 5e-3 & \multicolumn{1}{c|}{1e-3}& & \\
\cline{1-12}
\toprule[2pt]
\end{tabular}
\end{threeparttable}
}
\end{center}\label{table:hyperparam_PATTERN}
\end{table*}

\begin{table*}[!ht]
\caption{Hyperparameter Settings of all experiments on on SBM CLUSTER. $\#Param$ is the number of training parameters; $L$ is the number of GNN module layers; $C_*$ are channel dimensions in Figure \ref{fig:FOG}, \ref{fig:GCN}, \ref{fig:GCN_FOG}, \ref{fig:GAT}, \ref{fig:GAT_FOG}, \ref{fig:GatedGCN}, \ref{fig:GatedGCN_FOG}, \ref{fig:GIN}, \ref{fig:GIN_FOG} , \ref{fig:GraphSAGE}, and \ref{fig:GraphSAGE_FOG}; $Init$ $lr$ is the initial learning rate; $wd$ is the weight decay; $Patience$ is the decay patience; $Min$ $lr$ is the stopping learning rate. Learning rate reduce factor and batch size are set to $0.5$ and $128$ for all experiments, respectively.}
\begin{center}
\scalebox{0.75}{
\begin{threeparttable}
\begin{tabular}{l|c|cccccccl|cccc}
\toprule[2pt]
\multirow{2}{*}{Model} & \multirow{2}{*}{\#Param} & \multicolumn{8}{c|}{Hyperparameters} & \multicolumn{4}{c}{Learning}\\
\cline{3-14}
&&$L$&$C_{in}/C_{out}$&$C_{h1}$&$C_{h2}$&$C_p/C_q$&$C_{FC_1}$&$C_{FC_2}$&Other&Init lr&\multicolumn{1}{c|}{wd}& Patience & Min lr \\
\hline
FOG & 99,770 & 4 & 144 & 16 & 8 & - & 72 & 36 & - & 5e-3 & \multicolumn{1}{c|}{0} & \multirow{13}{*}{\shortstack[c]{5}} & \multirow{13}{*}{\shortstack[c]{1e-5}}\\
\cline{1-12}
GCN & 101,655 & 4 & 146 & - && - & 73 & 36 & - & 1e-3  &  \multicolumn{1}{c|}{1e-5} & & \\
\cline{1-12}
GCN+FOG & 101,830 & 4 & 160 & 12 & 6 & 80 & 80 & 40 & - &  1e-2 & \multicolumn{1}{c|}{0} & &\\
\cline{1-12}
\multirow{2}{*}{\shortstack[l]{GAT}} & \multirow{2}{*}{\shortstack[c]{110,700}} & \multirow{2}{*}{\shortstack[c]{4}} & \multirow{2}{*}{\shortstack[c]{152}} & \multirow{2}{*}{\shortstack[c]{-}} & \multirow{2}{*}{\shortstack[c]{-}} & \multirow{2}{*}{\shortstack[c]{-}} & \multirow{2}{*}{\shortstack[c]{76}} & \multirow{2}{*}{\shortstack[c]{38}} & $K=8$ for the $1^{st}$ to $3^{rd}$ module; & \multirow{2}{*}{\shortstack[c]{1e-3}}  & \multicolumn{1}{c|}{\multirow{2}{*}{\shortstack[c]{0}}}  &  & \\
&  &  &  &  &  & & &  & $K=1$ for the $4^{th}$ module; &  & \multicolumn{1}{c|}{} & &\\
\cline{1-12}
\multirow{2}{*}{\shortstack[l]{GAT+FOG}} & \multirow{2}{*}{\shortstack[c]{102,150}} & \multirow{2}{*}{\shortstack[c]{4}} & \multirow{2}{*}{\shortstack[c]{160}} & \multirow{2}{*}{\shortstack[c]{12}} &\multirow{2}{*}{\shortstack[c]{6}} & \multirow{2}{*}{\shortstack[c]{80}} & \multirow{2}{*}{\shortstack[c]{80}} & \multirow{2}{*}{\shortstack[c]{40}} & $K=8$ for the $1^{st}$ to $3^{rd}$ module; & \multirow{2}{*}{\shortstack[c]{1e-2}}  &  \multicolumn{1}{c|}{\multirow{2}{*}{\shortstack[c]{0}}}& & \\
&  &  &  &  &  & & &  & $K=1$ for the $4^{th}$ module; &  & \multicolumn{1}{c|}{} & &\\
\cline{1-12}

GatedGCN & 104,355 & 4 & 70 & - & - & - & 35 & 17 & - & 1e-3  & \multicolumn{1}{c|}{0}  & & \\
\cline{1-12}
GatedGCN+FOG & 102,374 & 4 & 64  & 8 & 4 & 32 & 32 & 16 & - & 5e-3  & \multicolumn{1}{c|}{1e-6} & &\\
\cline{1-12}
GIN & 100,884 & 4 & 110 & - & - & - & - & - & - & 1e-3  & \multicolumn{1}{c|}{0} & & \\
\cline{1-12}
GIN+FOG & 102,806 & 4 & 148  & 12 & 6 & 74 & - & - & - & 1e-2 & \multicolumn{1}{c|}{1e-6}& & \\
\cline{1-12}
GraphSAGE & 102,187 & 4 & 89 & - & - & - & 44 & 22 & - & 1e-3  & \multicolumn{1}{c|}{0} & & \\
\cline{1-12}
GraphSAGE+FOG & 96,171 & 4 & 98  & 9 & 4 & 49 & 49 & 24 & - & 1e-2 & \multicolumn{1}{c|}{1e-6}& & \\

\cline{1-12}
\toprule[2pt]
\end{tabular}
\end{threeparttable}
}
\end{center}\label{table:hyperparam_CLUSTER}
\end{table*}

\begin{table*}[!ht]
\caption{Hyperparameter Settings of all experiments on ZINC. $\#Param$ is the number of training parameters; $L$ is the number of GNN module layers; $C_*$ are channel dimensions in Figure \ref{fig:FOG}, \ref{fig:GCN}, \ref{fig:GCN_FOG}, \ref{fig:GAT}, \ref{fig:GAT_FOG}, \ref{fig:GatedGCN}, \ref{fig:GatedGCN_FOG}, \ref{fig:GIN}, \ref{fig:GIN_FOG} , \ref{fig:GraphSAGE}, and \ref{fig:GraphSAGE_FOG}; $Init$ $lr$ is the initial learning rate; $wd$ is the weight decay; $Patience$ is the decay patience; $Min$ $lr$ is the stopping learning rate. Learning rate reduce factor and batch size are set to $0.5$ and $128$ for all experiments, respectively.}
\begin{center}
\scalebox{0.74}{
\begin{threeparttable}
\begin{tabular}{l|c|cccccccl|cccc}
\toprule[2pt]
\multirow{2}{*}{Model} & \multirow{2}{*}{\#Param} & \multicolumn{8}{c|}{Hyperparameters} & \multicolumn{3}{c}{Learning}\\
\cline{3-14}
&&$L$&$C_{in}/C_{out}$&$C_{h1}$&$C_{h2}$&$C_p/C_q$&$C_{FC_1}$&$C_{FC_2}$&Other&Init lr &\multicolumn{1}{c|}{wd}& Patience & Min lr \\
\hline
FOG & 101,668 & 4 & 143 & 16 & 8 &- & 71 & 35 & Readout: mean & 1e-3 & \multicolumn{1}{c|}{0} & \multirow{23}{*}{\shortstack[c]{10}} & \multirow{23}{*}{\shortstack[c]{1e-5}}\\
\cline{1-12}
GCN & 103,077 & 4 & 145 & - & - & - & 72 & 36 & Readout: mean & 1e-3 & \multicolumn{1}{c|}{0} & &\\
\cline{1-12}
\multirow{4}{*}{\shortstack[l]{GCN+FOG}} & 102,809 & \multirow{4}{*}{\shortstack[c]{4}} & 158 & 12 & 6 & 79 & 79 & 39 & \multirow{4}{*}{\shortstack[l]{Readout: mean}} & \multirow{4}{*}{\shortstack[c]{1e-2}} & \multicolumn{1}{c|}{\multirow{4}{*}{\shortstack[c]{0}}}& &\\
& 77,278 &  & 138  & 11 & 5 & 69 & 69 & 34 & & &\multicolumn{1}{c|}{} & &\\
& 50,547 &  & 108  & 10 & 5 & 54 & 54 & 27 & & &\multicolumn{1}{c|}{} & &\\
& 25,847 &  & 76   & 8 & 4 & 38 & 38 & 19 & & & \multicolumn{1}{c|}{}& &\\
\cline{1-12}
\multirow{3}{*}{\shortstack[l]{GAT}} & \multirow{3}{*}{\shortstack[c]{102,385}} & \multirow{3}{*}{\shortstack[c]{4}} & \multirow{3}{*}{\shortstack[c]{144}} & \multirow{3}{*}{\shortstack[c]{-}} &\multirow{3}{*}{\shortstack[c]{-}} & \multirow{3}{*}{\shortstack[c]{-}} & \multirow{3}{*}{\shortstack[c]{72}} & \multirow{3}{*}{\shortstack[c]{36}} & $K=8$ for the $1^{st}$ to $3^{rd}$ module; & \multirow{3}{*}{\shortstack[c]{1e-3}} & \multicolumn{1}{c|}{\multirow{3}{*}{\shortstack[c]{0}}} & &\\
&  &  &  &  & & &  &  & $K=1$ for the $4^{th}$ module; &  & \multicolumn{1}{c|}{} & &\\
&  &  &  &  & & &  &  & Readout: mean &  & \multicolumn{1}{c|}{} & & \\
\cline{1-12}
\multirow{3}{*}{\shortstack[l]{GAT+FOG}} & \multirow{3}{*}{\shortstack[c]{105,305}} & \multirow{3}{*}{\shortstack[c]{4}} & \multirow{3}{*}{\shortstack[c]{160}} & \multirow{3}{*}{\shortstack[c]{12}} & \multirow{3}{*}{\shortstack[c]{6}} &\multirow{3}{*}{\shortstack[c]{80}} & \multirow{3}{*}{\shortstack[c]{80}} & \multirow{3}{*}{\shortstack[c]{40}} & $K=8$ for the $1^{st}$ to $3^{rd}$ module; & \multirow{3}{*}{\shortstack[c]{1e-2}} & \multicolumn{1}{c|}{\multirow{3}{*}{\shortstack[c]{1e-6}}} &  &\\
&  &  &  &  & & &  &  & $K=1$ for the $4^{th}$ module; &  & \multicolumn{1}{c|}{} & &\\
&  &  &  &  & & &  &  & Readout: mean &  & \multicolumn{1}{c|}{} & &\\
\cline{1-12}

GatedGCN & 105,735 & 4 & 70 & - & - & - & 35 & 17 & Readout: mean & 1e-3 & \multicolumn{1}{c|}{0} & &\\
\cline{1-12}
GatedGCN+FOG & 103,633 & 4 & 64  & 8 & 4 & 32 & 32 & 16 & Readout: mean & 1e-2 & \multicolumn{1}{c|}{1e-6}& & \\
\cline{1-12}

GatedGCN-E & 105,875 & 4 & 70 & - & - & - & 35 & 17 & Readout: mean & 1e-3 & \multicolumn{1}{c|}{0} & &\\
\cline{1-12}
\multirow{4}{*}{\shortstack[l]{GatedGCN-E+FOG}} & 103,761 & \multirow{4}{*}{\shortstack[c]{4}} & 64 & 8 & 4 & 32 & 32 & 16 & \multirow{4}{*}{\shortstack[l]{Readout: mean}} & \multirow{4}{*}{\shortstack[c]{5e-3}} & \multicolumn{1}{c|}{\multirow{4}{*}{\shortstack[c]{0}}} & &\\
& 79,165 &  & 56  & 7 & 3 & 28 & 28 & 14 & & &\multicolumn{1}{c|}{} & & \\
& 49,835 &  & 44  & 6 & 3 & 22 & 22 & 11 & & & \multicolumn{1}{c|}{}& & \\
& 26,909 &  & 32  & 5 & 2 & 16 & 16 & 8  & & &\multicolumn{1}{c|}{} & & \\

\cline{1-12}
GIN & 103,079 & 4 & 110 & - & - & - & - & - & Readout: sum & 1e-3 & \multicolumn{1}{c|}{0} & &\\
\cline{1-12}
GIN+FOG & 102,189 & 4 & 148 & 12 & 6 & 74 & - & - & Readout: sum & 5e-3 & \multicolumn{1}{c|}{1e-3} & &\\

\cline{1-12}
GraphSAGE & 94,977 & 4 & 90 & - & - & - & 45 & 22 & Readout: mean & 1e-3 & \multicolumn{1}{c|}{0} & &\\
\cline{1-12}
GraphSAGE+FOG & 94,477 & 4 & 96 & 9 & 4 & 48 & 48 & 24 & Readout: mean & 1e-2 & \multicolumn{1}{c|}{1e-6} & &\\

\hline
\toprule[2pt]
\end{tabular}
\end{threeparttable}
}
\end{center}\label{table:hyperparam_ZINC}
\end{table*}

\begin{table*}[!ht]
\caption{Hyperparameter Settings of all experiments on ZINC withot node feature. $\#Param$ is the number of training parameters; $L$ is the number of GNN module layers; $C_*$ are channel dimensions in Figure \ref{fig:FOG},  \ref{fig:GatedGCN}, \ref{fig:GatedGCN_FOG}, \ref{fig:GIN}, and \ref{fig:GIN_FOG}; $Init$ $lr$ is the initial learning rate; $wd$ is the weight decay; $Patience$ is the decay patience; $Min$ $lr$ is the stopping learning rate. Learning rate reduce factor and batch size are set to $0.5$ and $128$ for all experiments, respectively.}
\begin{center}
\scalebox{0.74}{
\begin{threeparttable}
\begin{tabular}{l|c|cccccccl|cccc}
\toprule[2pt]
\multirow{2}{*}{Model} & \multirow{2}{*}{\#Param} & \multicolumn{8}{c|}{Hyperparameters} & \multicolumn{3}{c}{Learning}\\
\cline{3-14}
&&$L$&$C_{in}/C_{out}$&$C_{h1}$&$C_{h2}$&$C_p/C_q$&$C_{FC_1}$&$C_{FC_2}$&Other&Init lr &\multicolumn{1}{c|}{wd}& Patience & Min lr \\
\hline
FOG & 101,668 & 4 & 143 & 16 & 8 &- & 71 & 35 & Readout: mean & 5e-4 & \multicolumn{1}{c|}{1e-3} & \multirow{5}{*}{\shortstack[c]{10}} & \multirow{5}{*}{\shortstack[c]{1e-5}}\\
\cline{1-12}

GatedGCN & 105,735 & 4 & 70 & - & - & - & 35 & 17 & Readout: mean & 1e-3 & \multicolumn{1}{c|}{0} & &\\
\cline{1-12}
GatedGCN+FOG & 103,633 & 4 & 64  & 8 & 4 & 32 & 32 & 16 & Readout: mean & 1e-3 & \multicolumn{1}{c|}{1e-6}& & \\
\cline{1-12}

\cline{1-12}
GIN & 103,079 & 4 & 110 & - & - & - & - & - & Readout: sum & 1e-3 & \multicolumn{1}{c|}{0} & &\\
\cline{1-12}
GIN+FOG & 102,189 & 4 & 148 & 12 & 6 & 74 & - & - & Readout: sum & 1e-2 & \multicolumn{1}{c|}{0} & &\\

\hline
\toprule[2pt]
\end{tabular}
\end{threeparttable}
}
\end{center}\label{table:hyperparam_ZINC_wo_nodeFeat}
\end{table*}

\begin{table*}[!ht]
\caption{Hyperparameter Settings of all experiments on on TSP. $\#Param$ is the number of training parameters; $L$ is the number of GNN module layers; $C_*$ are channel dimensions in Figure \ref{fig:FOG}, \ref{fig:GCN}, \ref{fig:GCN_FOG}, \ref{fig:GAT}, \ref{fig:GAT_FOG}, \ref{fig:GatedGCN}, \ref{fig:GatedGCN_FOG}, \ref{fig:GIN}, \ref{fig:GIN_FOG} , \ref{fig:GraphSAGE}, and \ref{fig:GraphSAGE_FOG}; $Init$ $lr$ is the initial learning rate; $wd$ is the weight decay; $Patience$ is the decay patience; $Min$ $lr$ is the stopping learning rate. Learning rate reduce factor and batch size are set to $0.5$ and $32$ for all experiments, respectively.}
\begin{center}
\scalebox{0.74}{
\begin{threeparttable}
\begin{tabular}{l|c|cccccccl|cccc}
\toprule[2pt]
\multirow{2}{*}{Model} & \multirow{2}{*}{\#Param} & \multicolumn{8}{c|}{Hyperparameters} & \multicolumn{4}{c}{Learning}\\
\cline{3-14}
&&$L$&$C_{in}/C_{out}$&$C_{h1}$&$C_{h2}$&$C_p/C_q$&$C_{FC_1}$&$C_{FC_2}$&Other&Init lr& \multicolumn{1}{c|}{wd} &Patience & Min lr \\
\hline
FOG & 96,386 & 4 & 120 & 15 & 7 & - & 120 & 60 & - & 1e-2 & \multicolumn{1}{c|}{1e-6} & \multirow{15}{*}{\shortstack[c]{10}} & \multirow{15}{*}{\shortstack[c]{1e-5}}\\
\cline{1-12}
GCN & 95,702 & 4 & 120 & - & - & - & 120 & 60 & - &  1e-3 &  \multicolumn{1}{c|}{0} & &\\
\cline{1-12}
GCN+FOG & 93,465 & 4 & 126 & 11 & 5 & 63 & 126 & 63 & - & 1e-2 & \multicolumn{1}{c|}{1e-6} & &\\
\cline{1-12}
\multirow{2}{*}{\shortstack[l]{GAT}} & \multirow{2}{*}{\shortstack[c]{96,182}} & \multirow{2}{*}{\shortstack[c]{4}} & \multirow{2}{*}{\shortstack[c]{120}} & \multirow{2}{*}{\shortstack[c]{-}} & \multirow{2}{*}{\shortstack[c]{-}} & \multirow{2}{*}{\shortstack[c]{-}} & \multirow{2}{*}{\shortstack[c]{120}} & \multirow{2}{*}{\shortstack[c]{60}} & $K=8$ for the $1^{st}$ to $3^{rd}$ module; & \multirow{2}{*}{\shortstack[c]{1e-3}}  &  \multicolumn{1}{c|}{\multirow{2}{*}{0}} & &\\
&  &  &  &  &  & & &  & $K=1$ for the $4^{th}$ module; &  & \multicolumn{1}{c|}{} & &\\
\cline{1-12}
\multirow{2}{*}{\shortstack[l]{GAT+FOG}} & \multirow{2}{*}{\shortstack[c]{96,350}} & \multirow{2}{*}{\shortstack[c]{4}} & \multirow{2}{*}{\shortstack[c]{128}} & \multirow{2}{*}{\shortstack[c]{11}} &\multirow{2}{*}{\shortstack[c]{5}} & \multirow{2}{*}{\shortstack[c]{64}} & \multirow{2}{*}{\shortstack[c]{128}} & \multirow{2}{*}{\shortstack[c]{64}} & $K=8$ for the $1^{st}$ to $3^{rd}$ module; & \multirow{2}{*}{\shortstack[c]{1e-2}}  & \multicolumn{1}{c|}{\multirow{2}{*}{{1e-6}}} & &\\
&  &  &  &  &  & & &  & $K=1$ for the $4^{th}$ module; &  & \multicolumn{1}{c|}{} & &\\
\cline{1-12}

GatedGCN & 97,858 & 4 & 65 & - & - & - & 65 & 32 & - & 1e-3   & \multicolumn{1}{c|}{0} & &\\
\cline{1-12}
GatedGCN+FOG & 95,456 & 4 & 60  & 7 & 3 & 30 & 60 & 30 & - & 1e-2 & \multicolumn{1}{c|}{1e-6} & &\\
\cline{1-12}
GatedGCN-E & 97,858 & 4 & 65 & - & - & - & 65 & 32 & - & 1e-3  &  \multicolumn{1}{c|}{0} & &\\
\cline{1-12}
GatedGCN-E+FOG & 95,456 & 4 & 60  & 7 & 3 & 30 & 60 & 30 & - & 1e-2 & \multicolumn{1}{c|}{1e-6} & &\\
\cline{1-12}
GIN & 99,002 & 4 & 73 & - & - & - & - & - & - & 1e-3  &  \multicolumn{1}{c|}{0} & & \\
\cline{1-12}
GIN+FOG & 94,046 & 4 & 80  & 8 & 4 & 40 & - & - & - & 1e-2  & \multicolumn{1}{c|}{1e-6} & &\\
\cline{1-12}
GraphSAGE & 99,263 & 4 & 82 & - & - & - & 82 & 41 & - & 1e-3  &  \multicolumn{1}{c|}{0} & & \\
\cline{1-12}
GraphSAGE+FOG & 97,007 & 4 & 90  & 9 & 4 & 45 & 90 & 45 & - & 5e-3  & \multicolumn{1}{c|}{1e-6} & &\\
\hline
\toprule[2pt]
\end{tabular}
\end{threeparttable}
}
\end{center}\label{table:hyperparam_TSP}
\end{table*}

\begin{table*}[!ht]
\caption{Hyperparameter Settings of all experiments on IMDB-MULTI. $\#Param$ is the number of training parameters; $L$ is the number of GNN module layers; $C_*$ are channel dimensions in Figure \ref{fig:FOG}, \ref{fig:GatedGCN}, \ref{fig:GatedGCN_FOG}, \ref{fig:GIN}, and \ref{fig:GIN_FOG}; $Init$ $lr$ is the initial learning rate; $wd$ is the weight decay; $Epochs$ is the total number of training epochs; $Step$ means the learning rate reduced after every fixed number of epochs; $Dropout$ is the dropout rate. Learning rate reduce factor and batch size are set to $0.5$ and $128$ for all experiments, respectively.}
\begin{center}
\scalebox{0.8}{
\begin{threeparttable}
\begin{tabular}{l|c|cccccccl|ccccc}
\toprule[2pt]
\multirow{2}{*}{Model} & \multirow{2}{*}{\#Param} & \multicolumn{8}{c|}{Hyperparameters} & \multicolumn{5}{c}{Learning}\\
\cline{3-15}
&&$L$&$C_{in}/C_{out}$&$C_{h1}$&$C_{h2}$&$C_p/C_q$&$C_{FC_1}$&$C_{FC_2}$&Other&Init lr& \multicolumn{1}{c|}{wd} & Epochs & Step & Dropout \\
\hline
FOG & 33017 & 4 & 81  & 12 & 6 & - & 40 & 20 & Readout: sum & 1e-2 & \multicolumn{1}{c|}{1e-6} &
\multirow{5}{*}{\shortstack[c]{350}} & \multirow{5}{*}{\shortstack[c]{50}} & 
\multirow{5}{*}{\shortstack[c]{0.5}}\\
\cline{1-12}
GatedGCN & 34663 & 4 & 40 & - & - & - & 20 & 10 & Readout: sum &  5e-3 & \multicolumn{1}{c|}{1e-6}  & & & \\
\cline{1-12}
GatedGCN+FOG & 33303 & 4 & 36 & 6 & 3 & 18 & 18 & 9 & Readout: sum & 1e-3  & \multicolumn{1}{c|}{1e-3} & & &   \\
\cline{1-12}
GIN & 35411 & 4 & 64 & - & - & - & - & - & Readout: sum & 1e-2 & \multicolumn{1}{c|}{0} & & &  \\
\cline{1-12}
GIN+FOG & 34805 & 4 & 86  & 9 & 4 & 43 & - & - & Readout: sum & 1e-2 & \multicolumn{1}{c|}{1e-6} & &   &  \\

\hline
\toprule[2pt]
\end{tabular}
\end{threeparttable}
}
\end{center}\label{table:hyperparam_IMDBMULTI}
\end{table*}

\begin{table*}[!ht]
\caption{Hyperparameter Settings of all experiments on miniImageNet, 5-way 1-shot task. $\#Param$ is the number of training parameters in the GNN part; $L$ is the number of GNN module layers; $C_*$ are channel dimensions in Figure \ref{fig:FOG}, \ref{fig:Few_shot_GNN}, and \ref{fig:Few_shot_GNN_FOG}; $Init$ $lr$ is the initial learning rate; $wd$ is the weight decay; $Epochs$ is the total number of training epochs; $Step$ means the learning rate reduced after every fixed number of epochs; Learning rate reduce factor and batch size are set to $0.5$ and $100$ for all experiments, respectively.}
\begin{center}
\scalebox{0.8}{
\begin{threeparttable}
\begin{tabular}{l|c|cccccccl|cccc}
\toprule[2pt]
\multirow{2}{*}{Model} & \multirow{2}{*}{\#Param} & \multicolumn{8}{c|}{Hyperparameters} & \multicolumn{4}{c}{Learning}\\
\cline{3-14}
&&$L$&$C_{in}^1$&$C_{in}^2$&$C_{in}^3$&$C_{h1}$&$C_{h2}$&$C_p/C_q$&$C_{A}$&Init lr& \multicolumn{1}{c|}{wd} & Epochs & Step \\
\hline
FOG & 312,282 & 3 & 133  & 181 & 229 & 9 & 4 & - & 48 & 5e-4 & \multicolumn{1}{c|}{0} &
\multirow{3}{*}{\shortstack[c]{80000}} & \multirow{3}{*}{\shortstack[c]{15000}}\\
\cline{1-12}
GNN & 335,994 & 3 & 133 & 181 & 229 & - & - & - & 48 &  1e-3 & \multicolumn{1}{c|}{1e-6}  & &  \\
\cline{1-12}
GNN+FOG & 323,760 & 3 & 133 & 181 & 229 & 6 & 3 & 24 & 48& 5e-3  & \multicolumn{1}{c|}{0} & &    \\

\hline
\toprule[2pt]
\end{tabular}
\end{threeparttable}
}
\end{center}\label{table:hyperparam_5w1s}
\end{table*}

\begin{table*}[!ht]
\caption{Hyperparameter Settings of all experiments on miniImageNet, 5-way 5-shot task. $\#Param$ is the number of training parameters in the GNN part; $L$ is the number of GNN module layers; $C_*$ are channel dimensions in Figure \ref{fig:FOG}, \ref{fig:Few_shot_GNN}, and \ref{fig:Few_shot_GNN_FOG}; $Init$ $lr$ is the initial learning rate; $wd$ is the weight decay; $Epochs$ is the total number of training epochs; $Step$ means the learning rate reduced after every fixed number of epochs; Learning rate reduce factor and batch size are set to $0.5$ and $40$ for all experiments, respectively.}
\begin{center}
\scalebox{0.8}{
\begin{threeparttable}
\begin{tabular}{l|c|cccccccl|cccc}
\toprule[2pt]
\multirow{2}{*}{Model} & \multirow{2}{*}{\#Param} & \multicolumn{8}{c|}{Hyperparameters} & \multicolumn{4}{c}{Learning}\\
\cline{3-14}
&&$L$&$C_{in}^1$&$C_{in}^2$&$C_{in}^3$&$C_{h1}$&$C_{h2}$&$C_p/C_q$&$C_{A}$&Init lr& \multicolumn{1}{c|}{wd} & Epochs & Step \\
\hline
FOG & 312,282 & 3 & 133  & 181 & 229 & 9 & 4 & - & 48 & 5e-3 & \multicolumn{1}{c|}{0} &
\multirow{3}{*}{\shortstack[c]{90000}} & \multirow{3}{*}{\shortstack[c]{15000}}\\
\cline{1-12}
GNN & 335,994 & 3 & 133 & 181 & 229 & - & - & - & 48 &  1e-3 & \multicolumn{1}{c|}{1e-6}  & &  \\
\cline{1-12}
GNN+FOG & 323,760 & 3 & 133 & 181 & 229 & 6 & 3 & 24 & 48& 5e-3  & \multicolumn{1}{c|}{1e-6} & &    \\

\hline
\toprule[2pt]
\end{tabular}
\end{threeparttable}
}
\end{center}\label{table:hyperparam_5w5s}
\end{table*}
\end{appendices}

\end{document}